\documentclass[journal]{IEEEtran}

\usepackage{lineno}
\usepackage[ruled,vlined,linesnumbered]{algorithm2e}
\usepackage{multirow}
\usepackage{makecell}

\usepackage{amsmath}

\usepackage{booktabs}
\usepackage[table]{xcolor}

\usepackage{xcolor}
\usepackage{hyperref}
\definecolor{sciblue}{RGB}{0,0,255}

\hypersetup{
    colorlinks=true,       
    linkcolor=sciblue,     
    citecolor=sciblue,     
    urlcolor=magenta,      
    breaklinks=true        
}
\usepackage{graphicx}

\usepackage{amssymb}

\begin{document}

\title{ScalLoc: Continuous Oblique Thermal UAV Geo-Localization with a 2D Satellite Map}

\author{Xiaoran~Zhang$^{\dagger}$,
        Yu~Liu$^{\dagger}$,
        Jinyu~Liang,
        Chen~Liu,
        and~Huaxin~Xiao$^{*}$%
\thanks{Xiaoran Zhang and Yu Liu contributed equally to this work.
\emph{(Corresponding author: Huaxin Xiao.)}}%
\thanks{Xiaoran Zhang, Yu Liu, Jinyu Liang, and Huaxin Xiao are with the
College of Systems Engineering, National University of Defense Technology,
Changsha 410073, China
(e-mail: \{zhangxiaoran, liuyu20, liangjinyu, xiaohuaxin\}@nudt.edu.cn).}%
\thanks{Chen Liu is with the Beijing Satellite Navigation Center,
Beijing 100094, China
(e-mail: liuchen\_2010\_rs@whu.edu.cn).}%
}

\markboth{Journal of \LaTeX\ Class Files,~Vol.~13, No.~9, September~2014}%
{Shell \MakeLowercase{\textit{et al.}}: Bare Demo of IEEEtran.cls for Journals}

\maketitle


\IEEEpeerreviewmaketitle

\begin{abstract}
Cross-modal Thermal Geo-localization (TG) provides an illumination-robust positioning solution for Unmanned Aerial Vehicles (UAVs) in Global Navigation Satellite System (GNSS)-denied environments. However, existing thermal geo-localization methods mainly focus on near-nadir observations or depend on map-side elevation priors, limiting their applicability to continuous oblique thermal localization with only geo-referenced 2D satellite maps. To address this problem, we propose ScalLoc, a scale-aligned local-to-map localization framework for continuous oblique thermal UAV geo-localization. ScalLoc follows a reconstruction--refinement--anchoring--propagation design: it reconstructs local window geometry from thermal sequences, refines camera poses using track-based pose optimization, adaptively anchors each window to the satellite map, and propagates validated states to recover frame-level global trajectories. To support systematic evaluation, we construct Thermal-Oblique, a real thermal UAV sequence dataset containing 216 urban and rural day--night trajectories and 28,112 thermal frames with frame-level annotations. Extensive experiments demonstrate that ScalLoc reduces the mean localization error from 112.69~m to 26.97~m compared with the VGGT-only configuration and achieves the lowest frame-weighted mean position error among the compared methods. These results demonstrate the feasibility of continuous oblique thermal geo-localization using only lightweight 2D satellite maps, without requiring digital surface models or preconstructed 3D maps. Code and dataset are available at \url{https://github.com/FloralHercules/ScalLoc}.
\end{abstract}

\section{Introduction}
\label{sec:introduction}

With the increasing demand for autonomous UAV inspection, mapping, and navigation, reliable global positioning has become essential for Unmanned Aerial Vehicles (UAVs)~\cite{shakhatreh2019unmanned,avola2024uav}. Most UAV platforms rely on a Global Navigation Satellite System (GNSS), but blockage, multipath propagation, jamming, and spoofing can substantially degrade its reliability~\cite{isik2020integrity,burbank2024detecting}. Visual Geo-localization (VG) provides a complementary solution by matching onboard images to a preloaded geo-referenced map~\cite{couturier2024review,avola2024uav}. However, visible-light imagery becomes unreliable at night and under weak illumination. Thermal Geo-localization (TG) therefore uses onboard thermal images for map matching, enabling localization across day--night and adverse-illumination conditions~\cite{xiao2023long,xiao2024sthn}. Beyond the sensing modality, the geometric representation of the reference map further determines which camera viewpoints can be localized.

\begin{figure}[!t]
  \centering
  \includegraphics[width=\linewidth]{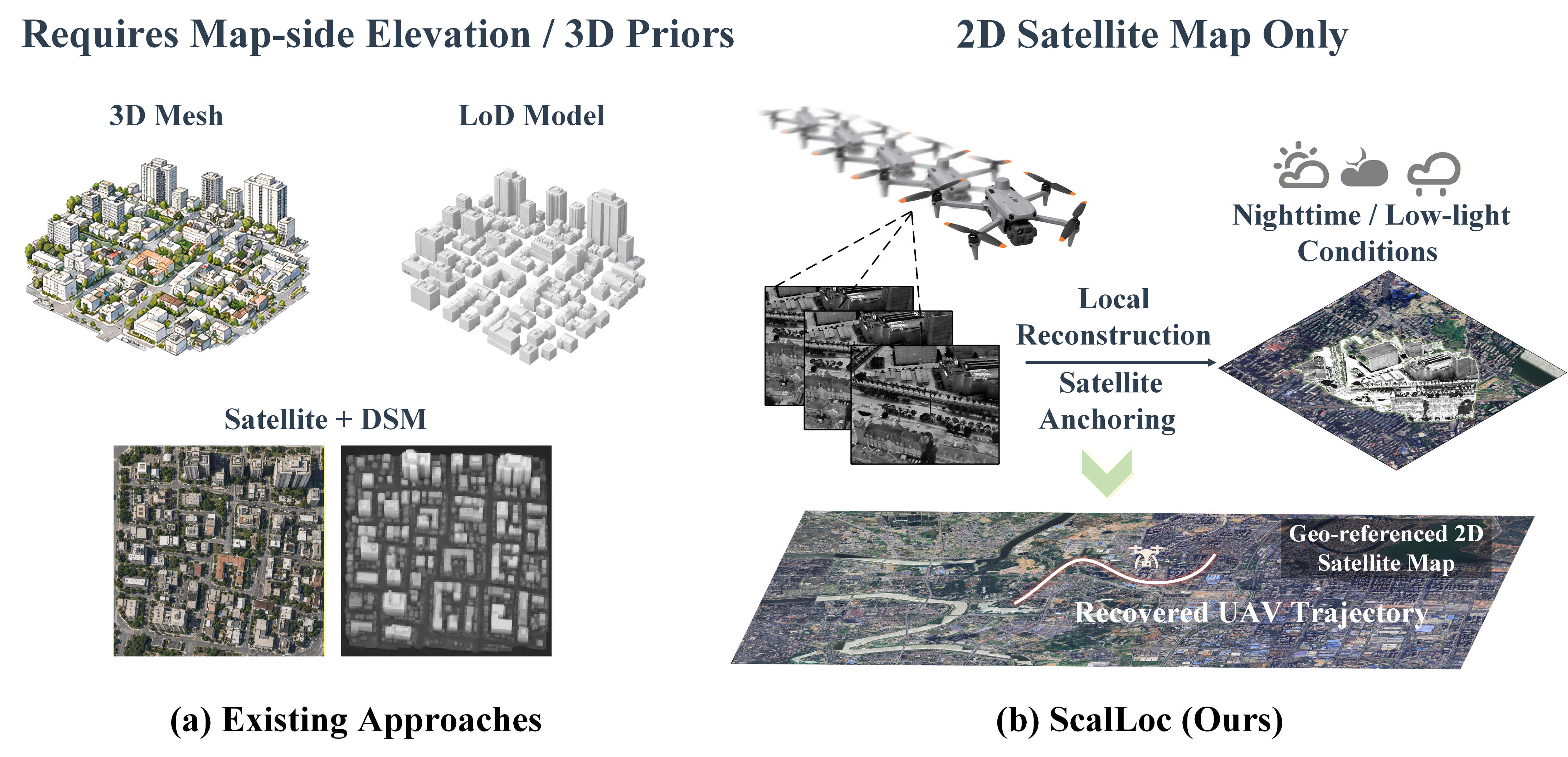}
    \caption{Comparison of UAV geo-localization frameworks with different reference-map priors. (a) Existing approaches using map-side elevation or 3D priors; (b) ScalLoc using only a geo-referenced 2D satellite map.}
    \label{Diagram}
\end{figure}

As illustrated in Fig.~\ref{Diagram}(a), existing UAV geo-localization methods for oblique viewpoints often require rich map-side geometric priors, such as geo-referenced 3D meshes, Level-of-Detail (LoD) city models, or satellite orthophotos with a Digital Surface Model (DSM). These priors support oblique localization through rendering-based matching, feature alignment, or explicit 2D-to-3D registration~\cite{wu2024uavd4l,zhu2024lodloc,dhaouadi2025ortholoc}. However, such elevation or 3D references are expensive to acquire, unavailable in many deployment regions, and difficult to maintain.

A more accessible alternative is a geo-referenced 2D satellite map, which provides horizontal map coordinates but no elevation. This lightweight representation is easy to obtain and maintain, but the lack of elevation makes oblique-view localization difficult because image appearance depends on scene height and viewing geometry. Existing satellite-only methods therefore typically rely on retrieval, planar alignment, or homography estimation, and are more naturally suited to near-nadir observations~\cite{zheng2020university,chen2024fpi}.

Motivated by this gap, we address continuous localization of oblique thermal UAV sequences using only geo-referenced 2D satellite maps. Without map-side 3D geometry such as DSMs, LoD models, or geo-referenced 3D scenes, this setting introduces two technical challenges and one benchmark gap:

\begin{itemize}

    \item \textbf{Height ambiguity under oblique views.} A 2D satellite map provides horizontal coordinates but no elevation. For near-nadir views, UAV and satellite images can often be approximated by planar alignment or homography estimation. Under oblique views, however, terrain and object heights introduce depth-dependent parallax, making purely 2D satellite--thermal matching insufficient for reliable pose recovery with a single homography~\cite{hartley2003multiple}.

    \item \textbf{Accumulated errors in recovered geometric poses.} Sequence reconstruction can recover locally coherent geometry and relative camera poses, but the recovered poses remain in a local coordinate frame and may accumulate scale, orientation, and pose errors over long horizons~\cite{wang2025vggt}.

    \item \textbf{Benchmark gap for continuous oblique thermal localization.} As summarized in Table~\ref{tab:uav_geoloc_datasets}, existing UAV geo-localization benchmarks mainly target visible-light imagery, near-nadir views, image-level retrieval, or settings with map-side 3D/elevation references. Existing thermal datasets provide valuable satellite--thermal data, but they rarely support continuous oblique-view localization with frame-level trajectories and only a 2D satellite map.
\end{itemize}


To address the above challenges, we propose ScalLoc, i.e., Scale-aligned Local-to-map Localization, for continuous oblique thermal UAV geo-localization using only a geo-referenced 2D satellite map. As shown in Fig.~\ref{Diagram}(b), ScalLoc repeatedly anchors locally reconstructed window geometry to the global map, providing recurring geographic correction without map-side elevation priors. Specifically, ScalLoc follows a reconstruction--refinement--anchoring--propagation design: it reconstructs local window geometry from thermal sequences, refines local camera poses using track-based pose optimization, estimates motion-adaptive local-to-map transformations through satellite anchoring, and propagates only validated states across adjacent windows to recover a frame-level global trajectory. To support systematic evaluation, we introduce Thermal-Oblique, a real thermal UAV sequence dataset with oblique views, frame-level trajectories, and corresponding geo-referenced 2D satellite maps.





The main contributions are summarized as follows:

\begin{itemize}
    \item We propose ScalLoc, a continuous oblique thermal UAV geo-localization framework that estimates frame-level global positions and headings using only geo-referenced 2D satellite maps, without map-side elevation or preconstructed 3D priors.

    \item We introduce Thermal-Oblique, a real thermal UAV sequence dataset containing 216 day--night trajectories and 28,112 thermal frames with diverse scenes, motion patterns, flight heights, camera pitch angles, and frame-level annotations.

    \item We conduct extensive experiments showing that ScalLoc achieves the lowest frame-weighted mean position error, reducing ME from 112.69~m to 26.97~m compared with the VGGT-only configuration.
\end{itemize}

The remainder of this paper is organized as follows.
Section~\ref{sec:related_work} reviews related work.
Section~\ref{sec:method} presents ScalLoc.
Section~\ref{sec:Experimental Setup} describes the dataset, metrics, baselines,
and implementation details. Section~\ref{sec:Results} reports the experimental
results. Section~\ref{sec:discussion} discusses findings and limitations, and
Section~\ref{sec:conclusion} concludes this work.

\begin{table*}[!t]
    \centering
    \caption{Comparison of representative UAV geo-localization datasets. ``Metric Position'' indicates support for metric position evaluation, while ``Trajectory-level Eval.'' indicates support for continuous trajectory-level localization with frame-level annotations.}
    
    \label{tab:uav_geoloc_datasets}
    \renewcommand{\arraystretch}{1}
    \setlength{\tabcolsep}{3.5pt}
    \resizebox{\textwidth}{!}{%
    \begin{tabular}{@{}lccccccc@{}}
        \toprule
        \textbf{Dataset}
        & \textbf{Year}
        & \textbf{UAV Modality}
        & \textbf{Reference Map}
        & \makecell{\textbf{Scale}\\\textbf{(images)}}
        & \textbf{Viewpoint}
        & \makecell{\textbf{Metric}\\\textbf{Position}}
        & \makecell{\textbf{Trajectory-level}\\\textbf{Eval.}}
        \\
        \midrule

        University-1652~\cite{zheng2020university}
        & 2020
        & Synthetic RGB
        & Satellite
        & 50.2K
        & Multi-view
        & $\times$
        & $\times$
        \\

        SUES-200~\cite{zhu2023sues}
        & 2023
        & Real RGB
        & Satellite
        & 24.1K
        & Multi-view
        & $\times$
        & $\times$
        \\

        UAVD4L~\cite{wu2024uavd4l}
        & 2024
        & Real RGB
        & 3D mesh + DSM
        & 1.6K
        & Near-nadir
        & \checkmark
        & $\times$
        \\

        UAVD4L-LoD~\cite{zhu2024lodloc}
        & 2024
        & Real RGB
        & LoD3.0 model
        & 3.8K
        & Near-nadir
        & \checkmark
        & $\times$
        \\

        AnyVisLoc~\cite{Ye_2026_CVPR}
        & 2025
        & Real RGB
        & Satellite + DSM
        & 18.0K
        & Multi-view
        & \checkmark
        & $\times$
        \\

        OrthoLoC~\cite{dhaouadi2025ortholoc}
        & 2025
        & Real RGB
        & Satellite + DSM
        & 16.4K
        & Nadir + oblique
        & \checkmark
        & $\times$
        \\

        MovingDrone~\cite{dhaouadi2026orthotrack}
        & 2026
        & Synthetic RGB
        & \makecell{Satellite + DSM + LiDAR + 3D/LoD}
        & 343K
        & Nadir + oblique
        & \checkmark
        & \checkmark
        \\

        \midrule

        Boson-nighttime~\cite{xiao2023long}
        & 2023
        & Real thermal
        & Satellite
        & 49.8K
        & Nadir
        & $\times$
        & $\times$
        \\

        Thermal-UAV~\cite{zhang2026scc}
        & 2026
        & Real thermal
        & Satellite + DSM
        & 11.9K
        & Nadir
        & \checkmark
        & $\times$
        \\

        \textbf{Thermal-Oblique (Ours)}
        & 2026
        & \textbf{Real thermal}
        & \textbf{Satellite}
        & \textbf{28.1K}
        & \textbf{Oblique}
        & \checkmark
        & \checkmark
        \\

        \bottomrule
    \end{tabular}%
    }
\end{table*}

\section{Related Work}
\label{sec:related_work}

Related work on continuous oblique thermal UAV geo-localization can be grouped into three directions: map-side geometric priors for oblique localization, 2D satellite maps for retrieval and alignment, and sequence reconstruction with relative geometry.

\subsection{Map-Side Geometric Priors for Oblique Localization}

Map-side 3D geometric priors provide explicit scene geometry for oblique UAV localization by linking onboard image observations with geo-referenced 3D structures. UAVD4L~\cite{wu2024uavd4l} provides 3D mesh and DSM data for 6-DoF localization, while LoD-Loc~\cite{zhu2024lodloc} replaces dense textured meshes with structured city models and aligns image structures with projected LoD primitives. These methods handle large viewpoint changes using explicit scene geometry and height information, but they require costly 3D/LoD map construction and are sensitive to incomplete, outdated, or inaccurate map-side geometry.

Another line of work uses satellite image blocks and DSM as a lighter 2.5D reference, where satellite imagery provides appearance cues and DSM supplies elevation for metric pose estimation. OrthoLoC~\cite{dhaouadi2025ortholoc} estimates UAV 6-DoF poses by matching UAV images with satellite image blocks, lifting matched map pixels to 3D using DSM heights, and refining the pose through homography-based view alignment. OrthoTrack~\cite{dhaouadi2026orthotrack} extends elevation-assisted satellite localization to continuous trajectory estimation by anchoring keyframes to orthophotos and propagating correspondences with optical flow. Although easier to obtain than full 3D meshes or LoD models, these methods still require reliable elevation data and accurate map alignment, making them less suitable for 2D-map-only deployment.

\subsection{2D Satellite Maps for Retrieval and Alignment}

Geo-referenced 2D satellite maps are widely used in UAV geo-localization because they are easier to obtain and maintain than 3D maps or DSM-assisted references. University-1652~\cite{zheng2020university} and SUES-200~\cite{zhu2023sues} support retrieval-based UAV-to-satellite localization, where satellite candidates are retrieved for a given UAV query. Following this paradigm, later methods improve candidate search with stronger transferable representations, such as AnyLoc~\cite{keetha2023anyloc}, which combines self-supervised DINOv2 features with VLAD aggregation for zero-shot place retrieval. However, retrieval mainly provides a coarse candidate region and cannot directly recover metric camera pose or continuous trajectory.

Some methods further formulate fine geo-localization as local refinement within a retrieved or predefined search region. OS-FPI~\cite{chen2024fpi} couples candidate classification with offset regression to predict fine-grained spatial displacement, while map-retrieval-and-alignment methods~\cite{he2024leveraging} refine the candidate region through local geometric alignment. These methods improve position accuracy under near-nadir or weak-perspective settings, but their 2D offset or regression formulation is less suited to oblique views where the camera pose is affected by height-dependent parallax.

Recent dense matchers, including RoMa~\cite{edstedt2024roma}, XoFTR~\cite{tuzcuouglu2024xoftr}, and MINIMA~\cite{ren2025minima}, improve cross-view or cross-modal correspondences with stronger local matching representations. However, under oblique thermal views, matched 2D correspondences alone cannot reliably determine the camera pose from a satellite map without elevation information, because height-dependent parallax cannot be represented by a single planar homography. Therefore, retrieval, regression, or matching-based alignment remains insufficient for stable continuous trajectory-level localization with only 2D satellite maps.

\subsection{Sequence Reconstruction and Relative Geometry}

Image sequences provide temporal continuity that is unavailable in single-frame localization. Traditional SLAM and visual odometry methods, such as ORB-SLAM3~\cite{campos2021orb} and DROID-SLAM~\cite{teed2021droid}, estimate relative camera motion from consecutive observations through feature tracking, optical-flow-based correspondence modeling, or recurrent optimization. These methods can produce temporally coherent camera trajectories, but their estimates are primarily relative and depend on successful tracking over long sequences. In low-texture thermal imagery or under large viewpoint changes, tracking uncertainty can accumulate and gradually affect the recovered trajectory.

Recent feed-forward 3D vision models provide another route to sequence geometry. DUSt3R~\cite{wang2024dust3r} predicts point maps from image pairs, VGGT~\cite{wang2025vggt} jointly estimates camera parameters, depth, point maps, and tracks from multiple views, and SLAM3R~\cite{liu2025slam3r} extends point-map prediction to video fragments through progressive local registration. These models reduce the dependence on hand-crafted feature pipelines and provide useful local geometry for UAV image sequences. However, their outputs are still reconstructed in a local coordinate system and do not directly provide the local-to-global transformation.

Therefore, both traditional sequence estimation and feed-forward reconstruction can provide useful local geometry and relative camera poses, but they cannot by themselves recover global UTM positions or absolute headings. In particular, the recovered poses lack absolute geographic scale and map alignment, and local pose estimates can still accumulate scale, orientation, and pose errors over long trajectories.

\textbf{Limitations \& Motivation:} Existing methods and benchmarks have not fully addressed continuous oblique thermal UAV geo-localization with only a geo-referenced 2D satellite map. Elevation- or 3D-prior-based approaches handle oblique viewpoints but require costly DSMs, LoD models, or 3D meshes. Satellite-only retrieval and alignment methods are lightweight but mainly provide coarse regions, 2D offsets, or homography-based alignment, which is insufficient for oblique pose recovery. Sequence reconstruction provides local geometry but lacks absolute geographic scale and map alignment, and can accumulate long-horizon drift. Meanwhile, existing datasets rarely cover continuous oblique thermal trajectories under 2D-map-only conditions. 

To address these limitations, we introduce \textbf{Thermal-Oblique}, a real thermal UAV sequence benchmark designed for continuous oblique geo-localization. The dataset contains oblique thermal trajectories captured across diverse scenes, motion patterns, flight heights, camera pitch angles, and day--night conditions, with frame-level trajectory annotations, geo-referenced 2D satellite maps, and corresponding DSM data for evaluating elevation-assisted baselines. We further propose \textbf{ScalLoc}, a continuous thermal UAV geo-localization framework that reconstructs local sequence geometry, adaptively anchors each window to the satellite map, and propagates only validated corrections across windows. Unlike elevation- or 3D-prior-based methods, ScalLoc uses only the 2D satellite map during localization, connecting relative sequence reconstruction with absolute satellite-map localization without relying on map-side elevation or preconstructed 3D scenes.
\section{Method}
\label{sec:method}

\begin{figure*}[!t]
    \centering
    \includegraphics[width=0.9\textwidth]{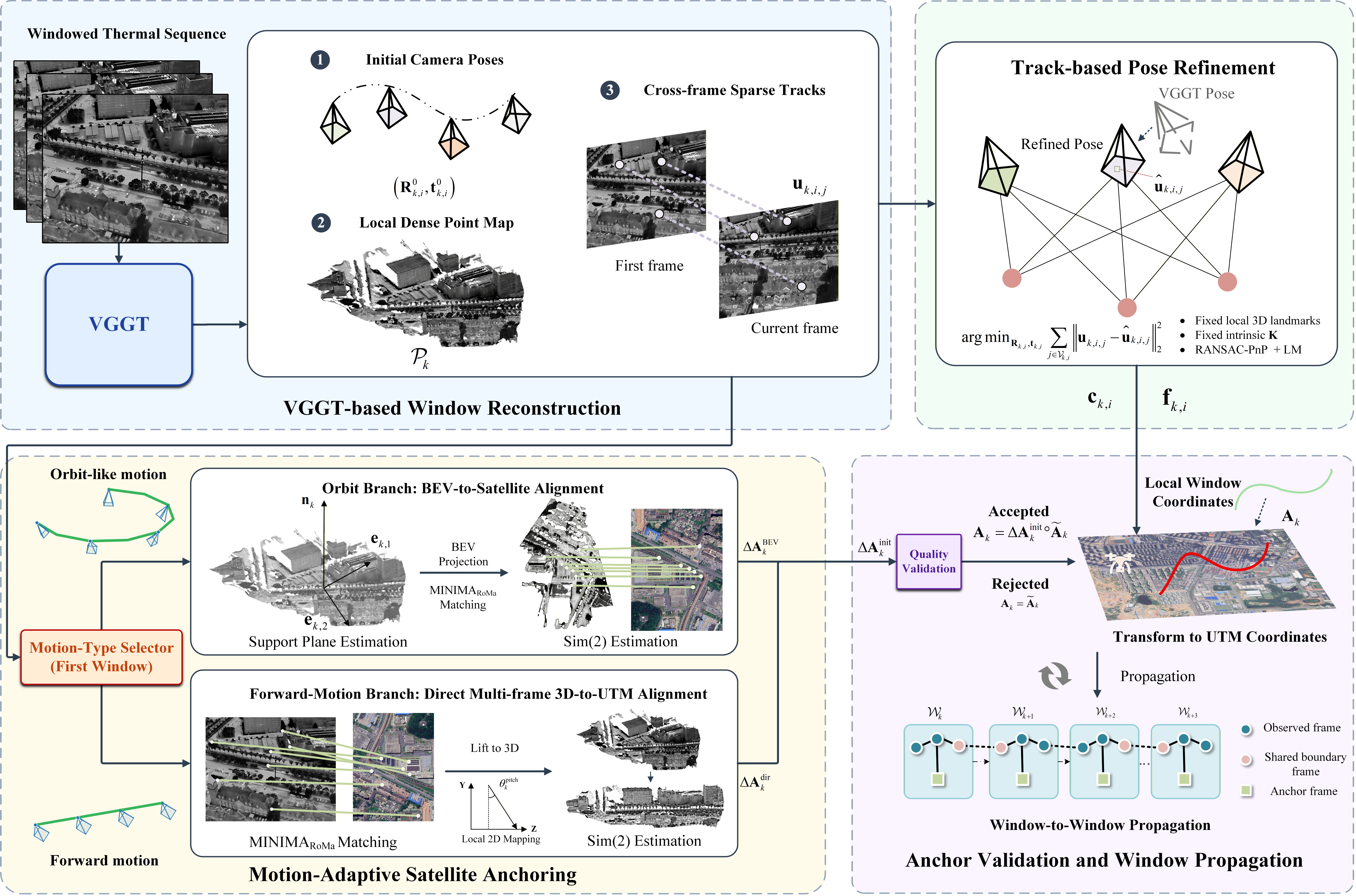}
    \caption{Overview of ScalLoc. The framework performs VGGT-based window reconstruction, track-based pose refinement, motion-adaptive satellite anchoring, anchor validation, and window-to-window propagation to recover frame-level UTM positions and headings from oblique thermal sequences.}
    \label{fig:scalloc_framework}
\end{figure*}

\subsection{Overview}

Given a thermal UAV image sequence and a geo-referenced satellite map, ScalLoc estimates the global UAV trajectory in UTM coordinates. As shown in Fig.~\ref{fig:scalloc_framework}, ScalLoc follows a reconstruction--refinement--anchoring--propagation pipeline. It first applies windowed VGGT reconstruction to obtain local dense point maps and initial camera poses, and then performs fixed-intrinsic track-based pose refinement to improve local camera geometry. A motion-adaptive selector infers the trajectory pattern from the first reconstructed window and chooses the satellite correction branch accordingly: orbit-like trajectories use BEV-to-satellite alignment, whereas forward-moving trajectories use direct multi-frame 3D-to-UTM alignment. The estimated residual correction is validated according to matching quality. Accepted corrections update the current local-to-map transformation, while rejected corrections retain the propagated prior. The validated state is finally propagated across adjacent windows to recover the global frame-level UAV positions and headings.

\subsection{Track-Based Window Pose Refinement}
\label{sec:track_optimized_window_reconstruction}

Given a thermal UAV sequence, we decompose it into short temporal windows to
obtain locally consistent geometry. Let $\mathcal{I}=\{I_t\}_{t=1}^{N}$ denote
the input sequence, where $t$ is the global frame index and $N$ is the sequence
length. Each window contains $L$ frames. For window $k$, $i$ denotes the local
frame index, and $\tau_{k,i}$ maps it to the corresponding global index. We
denote the $k$-th temporal window by $\mathcal{W}_k$, where adjacent windows
share one boundary frame:
\begin{equation}
\mathcal{W}_k
=
\{I_{\tau_{k,i}}\}_{i=1}^{L},
\qquad
\tau_{k,1}=\tau_{k-1,L}\quad (k>1).
\end{equation}

For each window, VGGT reconstructs a dense point map and initial camera poses in
a window-local coordinate frame $\mathcal{F}_k$. We denote the dense point map by $\mathcal{P}_k$. It consists of $N_k$ points in the window-local coordinate frame $\mathcal{F}_k$:

\begin{equation}
\mathcal{P}_k
=
\left\{
\mathbf{X}_{k,n}\in\mathbb{R}^{3}
\right\}_{n=1}^{N_k}.
\end{equation}

For frame $i$ in $\mathcal{W}_k$, let
$\mathbf{R}^{0}_{k,i}\in\mathrm{SO}(3)$ and
$\mathbf{t}^{0}_{k,i}\in\mathbb{R}^{3}$ denote the VGGT-predicted initial
extrinsics. We use the convention that these extrinsics transform points from
$\mathcal{F}_k$ to the camera frame:
\begin{equation}
\mathbf{x}^{\mathrm{cam}}_{k,i,n}
=
\mathbf{R}^{0}_{k,i}\mathbf{X}_{k,n}
+
\mathbf{t}^{0}_{k,i}.
\end{equation}

The initial camera center of frame $i$, expressed in $\mathcal{F}_k$, is:

\begin{equation}
\mathbf{c}^{0}_{k,i}
=
-\left(\mathbf{R}^{0}_{k,i}\right)^{\top}\mathbf{t}^{0}_{k,i}.
\end{equation}

VGGT also predicts sparse cross-frame tracks within each temporal window to
refine the initial camera geometry. Each track $j$ starts from a query point in
the first frame of $\mathcal{W}_k$ and is propagated to the visible frames. We
denote its tracked 2D pixel coordinate in frame $i$ as
$\mathbf{u}_{k,i,j}\in\mathbb{R}^{2}$, and initialize the corresponding sparse
3D landmark from the first-frame VGGT point map as
$\mathbf{Z}_{k,j}\in\mathbb{R}^{3}$. With the fixed calibrated intrinsic matrix
$\mathbf{K}\in\mathbb{R}^{3\times3}$, the projection of
$\mathbf{Z}_{k,j}$ under pose $\{\mathbf{R}_{k,i},\mathbf{t}_{k,i}\}$ is:

\begin{equation}
\hat{\mathbf{u}}_{k,i,j}
=
\pi
\left(
\mathbf{K}
\left(
\mathbf{R}_{k,i}\mathbf{Z}_{k,j}
+
\mathbf{t}_{k,i}
\right)
\right),
\end{equation}
where $\pi(\cdot)$ denotes the perspective projection to 2D pixel coordinates.

Let $\mathcal{V}_{k,i}$ denote the set of valid track indices observed in frame $i$ of window $\mathcal{W}_k$. The track-based pose refinement fixes the first camera pose to define the local reference frame and remove the coordinate ambiguity. The remaining poses are initialized from VGGT and refined frame by frame. For each frame $i>1$ with sufficient valid 2D--3D observations, RANSAC-based Perspective-$n$-Point (PnP) estimates a candidate refined pose by reducing the reprojection error:

\begin{equation}
\left(
\mathbf{R}^{\star}_{k,i},
\mathbf{t}^{\star}_{k,i}
\right)
=
\arg\min_{\mathbf{R}_{k,i},\mathbf{t}_{k,i}}
\sum_{j\in\mathcal{V}_{k,i}}
\left\|
\mathbf{u}_{k,i,j}
-
\hat{\mathbf{u}}_{k,i,j}
\right\|_2^2 .
\end{equation}

The PnP solution is further refined by Levenberg--Marquardt optimization when
it has sufficient inlier support. If PnP fails, has too few inliers, or the
refinement is unreliable, the pose falls back to the original VGGT prediction:
\begin{equation}
\left(
\mathbf{R}_{k,i},
\mathbf{t}_{k,i}
\right)
=
\begin{cases}
\left(
\mathbf{R}^{\star}_{k,i},
\mathbf{t}^{\star}_{k,i}
\right), & \text{if the refinement is accepted},\\
\left(
\mathbf{R}^{0}_{k,i},
\mathbf{t}^{0}_{k,i}
\right), & \text{otherwise}.
\end{cases}
\end{equation}

For the fixed reference frame, we keep $\mathbf{R}_{k,1}=\mathbf{R}^{0}_{k,1}$ and $\mathbf{t}_{k,1}=\mathbf{t}^{0}_{k,1}$. During refinement, the sparse landmarks remain fixed and serve as 3D references initialized from the first-frame VGGT geometry. Unlike conventional joint bundle adjustment, only the camera poses are refined using track-based PnP and LM optimization. This conservative strategy improves frames supported by reliable track geometry while retaining the original VGGT poses for unreliable frames. The refined camera center, expressed in $\mathcal{F}_k$, is computed as:

\begin{equation}
\mathbf{c}_{k,i}
=
-\mathbf{R}_{k,i}^{\top}\mathbf{t}_{k,i}.
\end{equation}

The refined extrinsics provide local camera orientations and camera centers in
$\mathcal{F}_k$. Together with the dense point map, they serve as the local
geometric input to the satellite anchoring stage.

\subsection{Motion-Adaptive Satellite Anchoring}
\label{sec:motion_adaptive_state_machine}

After window reconstruction, $\mathcal{W}_k$ provides the dense point
map $\mathcal{P}_k$, camera extrinsics
$\{\mathbf{R}_{k,i},\mathbf{t}_{k,i}\}$, and camera centers
$\{\mathbf{c}_{k,i}\}$. Let $\mathcal{M}$ denote the geo-referenced satellite
map, and let $\mathcal{G}\subset\mathbb{R}^{2}$ denote the UTM map plane. Given
the propagated pre-anchor local-to-map prior $\tilde{\mathbf{A}}_k$, satellite anchoring
estimates a residual correction that aligns the current window with
$\mathcal{M}$.

A motion-adaptive state machine selects the anchoring branch from the first reconstructed window. We compare the estimated local motion direction with the UAV forward direction: approximately orthogonal directions indicate orbit-like motion, while otherwise the trajectory is treated as forward-moving. The initial anchoring branch is then selected as:
\begin{equation}
\Delta\mathbf{A}^{\mathrm{init}}_k
=
\begin{cases}
\Delta\mathbf{A}^{\mathrm{BEV}}_k,
& \text{orbit-like trajectories}, \\
\Delta\mathbf{A}^{\mathrm{dir}}_k,
& \text{forward-moving trajectories},
\end{cases}
\end{equation}
where $\Delta\mathbf{A}^{\mathrm{BEV}}_k$ and $\Delta\mathbf{A}^{\mathrm{dir}}_k$ are the residual corrections estimated by the BEV and direct branches, respectively.

\subsubsection{Orbit Branch: BEV-to-Satellite Alignment}
\label{sec:bev_to_satellite_alignment}

For orbit-like trajectories, camera motion is often tangential to the viewing
direction. Direct oblique-image matching can therefore become unreliable. The
BEV branch converts the local point map into a ground-aligned orthophoto and
matches it with the satellite map.

We first estimate a dominant support plane from the VGGT point map using RANSAC.
The plane parameters maximize the number of inlier 3D points:
\begin{equation}
(\mathbf{n}_k,d_k)
=
\arg\max_{\|\mathbf{n}\|_2=1,d}
\sum_{\mathbf{X}_{k,n}\in\mathcal{P}_k}
\mathbb{I}
\left(
\left|
\mathbf{n}^{\top}\mathbf{X}_{k,n}
+d
\right|
<
\epsilon_k
\right),
\end{equation}
where $\mathbb{I}(\cdot)$ is the indicator function, $\mathbf{n}_k$ is the plane
normal, $d_k$ is the plane offset, and $\epsilon_k$ is the inlier threshold.
The plane is refined by SVD using the RANSAC inliers. We set $\mathbf{o}_k$ as
the inlier centroid and orient $\mathbf{n}_k$ toward the camera centers. Two
orthonormal in-plane axes, $\mathbf{e}_{k,1}$ and $\mathbf{e}_{k,2}$, are then
constructed, where $\mathbf{e}_{k,1}$ follows the projected camera-motion
direction and $\mathbf{e}_{k,2}=\mathbf{n}_k\times\mathbf{e}_{k,1}$.

For any 3D point $\mathbf{r}\in\mathbb{R}^{3}$ in the window-local coordinate
frame $\mathcal{F}_k$, we define its BEV projection as:
\begin{equation}
\Pi^{\mathrm{BEV}}_k(\mathbf{r})
=
\begin{bmatrix}
(\mathbf{r}-\mathbf{o}_k)^{\top}\mathbf{e}_{k,1} \\
(\mathbf{r}-\mathbf{o}_k)^{\top}\mathbf{e}_{k,2}
\end{bmatrix}.
\end{equation}

The dense points are then projected onto the BEV plane:

\begin{equation}
\mathbf{y}^{\mathrm{BEV}}_{k,n}
=
\Pi^{\mathrm{BEV}}_k(\mathbf{X}_{k,n}),
\end{equation}
where $\mathbf{y}^{\mathrm{BEV}}_{k,n}\in\mathbb{R}^{2}$ denotes the BEV
coordinate of the $n$-th dense point.

The local points are rasterized into a BEV orthophoto $\mathbf{B}_k$ with a
valid mask. A satellite crop $\mathbf{T}^{\mathrm{BEV}}_k$ is extracted from
$\mathcal{M}$ around the prior-mapped BEV footprint under
$\tilde{\mathbf{A}}_k$ with a fixed search margin. MINIMA$_{\mathrm{RoMa}}$~\cite{ren2025minima,edstedt2024roma} matches the BEV orthophoto with the
satellite crop:

\begin{equation}
\mathcal{C}^{\mathrm{BEV}}_k
=
\operatorname{Match}
\left(
\mathbf{B}_k,
\mathbf{T}^{\mathrm{BEV}}_k
\right),
\end{equation}
where $\operatorname{Match}(\cdot)$ denotes MINIMA$_{\mathrm{RoMa}}$ matching
followed by valid-mask filtering. For each match
$l\in\mathcal{C}^{\mathrm{BEV}}_k$, let
$\mathbf{y}^{\mathrm{BEV}}_{k,l}$ denote the local BEV coordinate and
$\mathbf{g}^{\mathrm{sat}}_{k,l}\in\mathcal{G}$ denote the matched satellite
UTM coordinate. Its prior-mapped coordinate $\tilde{\mathbf{g}}^{\mathrm{BEV}}_{k,l}$ is:

\begin{equation}
\tilde{\mathbf{g}}^{\mathrm{BEV}}_{k,l}
=
\tilde{\mathbf{A}}_k(\mathbf{y}^{\mathrm{BEV}}_{k,l}).
\end{equation}

The BEV residual is estimated by RANSAC-based $\mathrm{Sim}(2)$ fitting from the
prior-mapped BEV coordinates to the matched satellite UTM coordinates:
\begin{equation}
\Delta\mathbf{A}^{\mathrm{BEV}}_k
=
\arg\min_{\Delta\mathbf{A}}
\sum_{l\in\mathcal{C}^{\mathrm{BEV}}_k}
\left\|
\mathbf{g}^{\mathrm{sat}}_{k,l}
-
\Delta\mathbf{A}
\left(
\tilde{\mathbf{g}}^{\mathrm{BEV}}_{k,l}
\right)
\right\|_2^2 .
\end{equation}

The BEV branch returns $\Delta\mathbf{A}^{\mathrm{BEV}}_k$ to the state machine.

\subsubsection{Forward-Motion Branch: Direct Multi-frame 3D-to-UTM Alignment}
\label{sec:direct_multiframe_alignment}

For forward-moving trajectories, the direct branch retains the original oblique
UAV observations. It constructs local-to-map correspondences by matching
selected thermal frames directly with satellite tiles and lifting the matched
UAV pixels to the VGGT point map. Unlike the BEV branch, it does not rasterize
the reconstructed window into a ground orthophoto.

Let $\mathcal{Q}_k$ be the selected keyframe set in window $\mathcal{W}_k$. For
each keyframe $i\in\mathcal{Q}_k$, a yaw-aligned satellite tile
$\mathbf{T}^{\mathrm{dir}}_{k,i}$ is extracted from $\mathcal{M}$ around the
ground location predicted by the propagated prior, using the onboard keyframe
pitch and relative height. The direct branch then matches the thermal frame with the satellite tile:

\begin{equation}
\mathcal{C}^{\mathrm{img}}_{k,i}
=
\operatorname{Match}
\left(
I_{\tau_{k,i}},
\mathbf{T}^{\mathrm{dir}}_{k,i}
\right)
=
\left\{
(\mathbf{q}^{\mathrm{uav}}_{k,i,l},
\mathbf{q}^{\mathrm{sat}}_{k,i,l},
w_{k,i,l})
\right\}_{l},
\end{equation}
where
$\mathbf{q}^{\mathrm{uav}}_{k,i,l}$ and
$\mathbf{q}^{\mathrm{sat}}_{k,i,l}$ are the matched UAV and satellite pixels,
respectively, and $w_{k,i,l}$ is the matching confidence. The satellite pixel is
converted to a UTM coordinate using the geo-reference of
$\mathbf{T}^{\mathrm{dir}}_{k,i}$:

\begin{equation}
\mathbf{g}^{\mathrm{sat}}_{k,i,l}
=
\Gamma_{k,i}
\left(
\mathbf{q}^{\mathrm{sat}}_{k,i,l}
\right),
\end{equation}
where $\Gamma_{k,i}:\mathbb{R}^{2}\rightarrow\mathcal{G}$ denotes the
tile-to-UTM mapping.

The matched UAV pixel is lifted to the VGGT point map of the same frame:
\begin{equation}
\mathbf{X}_{k,i,l}
=
\Lambda_{k,i}
\left(
\mathbf{q}^{\mathrm{uav}}_{k,i,l}
\right),
\end{equation}
where $\Lambda_{k,i}(\cdot)$ maps an image pixel to its corresponding 3D point
in the VGGT point map of frame $i$. The lifted point
$\mathbf{X}_{k,i,l}\in\mathbb{R}^{3}$ lies in the window-local coordinate frame
$\mathcal{F}_k$.

Since the satellite side provides a 2D UTM coordinate while the UAV side
provides a local 3D point, the direct branch converts each lifted VGGT point
into a local 2D source coordinate. Let
$\mathbf{X}_{k,i,l}=(X_{k,i,l},Y_{k,i,l},Z_{k,i,l})^{\top}$. The direct
3D-to-2D mapping is:
\begin{equation}
\mathbf{y}^{\mathrm{dir}}_{k,i,l}
=
\Pi^{\mathrm{dir}}_k
\left(
\mathbf{X}_{k,i,l}
\right)
=
\begin{bmatrix}
X_{k,i,l} \\
-\sin\theta^{\mathrm{pitch}}_k Y_{k,i,l}
+
\cos\theta^{\mathrm{pitch}}_k Z_{k,i,l}
\end{bmatrix},
\end{equation}
where $\Pi^{\mathrm{dir}}_k(\cdot)$ denotes the direct-branch 3D-to-2D mapping,
and $\theta^{\mathrm{pitch}}_k$ is the window-level prior pitch angle. This
mapping keeps the local $X$ coordinate and uses the prior pitch to convert the
$(Y,Z)$ components into the second 2D coordinate, without constructing a BEV
support plane.

Each valid match therefore provides a weighted correspondence
$(\mathbf{y}^{\mathrm{dir}}_{k,i,l},
\mathbf{g}^{\mathrm{sat}}_{k,i,l},w_{k,i,l})$ between a local direct coordinate
and a satellite UTM coordinate. All valid matches from the selected keyframes
are aggregated into the index set $\mathcal{C}^{\mathrm{dir}}_k$. Before
residual estimation, the propagated prior maps each local source coordinate to
its prior UTM coordinate:

\begin{equation}
\tilde{\mathbf{g}}^{\mathrm{dir}}_{k,i,l}
=
\tilde{\mathbf{A}}_k
\left(
\mathbf{y}^{\mathrm{dir}}_{k,i,l}
\right).
\end{equation}

The direct residual is estimated by RANSAC-based $\mathrm{Sim}(2)$ fitting from
the prior-mapped direct coordinates to the matched satellite UTM coordinates.
The final fitting objective can be written as:

\begin{equation}
\Delta\mathbf{A}^{\mathrm{dir}}_k
=
\arg\min_{\Delta\mathbf{A}}
\sum_{(i,l)\in\mathcal{C}^{\mathrm{dir}}_k}
w_{k,i,l}
\left\|
\mathbf{g}^{\mathrm{sat}}_{k,i,l}
-
\Delta\mathbf{A}
\left(
\tilde{\mathbf{g}}^{\mathrm{dir}}_{k,i,l}
\right)
\right\|_2^2 .
\end{equation}

The direct branch returns $\Delta\mathbf{A}^{\mathrm{dir}}_k$ to the state machine.

\subsection{Anchor Validation and Window Propagation}
\label{sec:anchor_validation_window_propagation}

After the motion-adaptive anchoring stage, the system obtains an initial
residual correction $\Delta\mathbf{A}^{\mathrm{init}}_k$ for the current window.
This residual is validated according to matching quality, including valid-match
count, $\mathrm{Sim}(2)$ inlier support, inlier ratio, residual error, and
spatial coverage. Residuals with weak matching support or large fitting error
are rejected.

For forward-moving trajectories, the direct branch is the primary correction
path. If the direct residual fails the quality validation or anchor rejection
occurs repeatedly, the state machine may activate two rescue strategies. The
first is BEV-guided direct retry, which uses the BEV alignment result to guide a
new direct matching attempt. The second is BEV rescue, which falls back to the
BEV residual when the direct branch remains unreliable. A rescue residual is
used only if it also passes the same quality validation. For notational
simplicity, we still denote the residual retained by the state machine as
$\Delta\mathbf{A}^{\mathrm{init}}_k$.

Let $a_k\in\{0,1\}$ indicate whether the retained residual is accepted, and let
$\mathbf{A}_k$ denote the final local-to-map transformation of
$\mathcal{W}_k$. It maps local 2D source coordinates to the UTM map plane and is
defined as:

\begin{equation}
\mathbf{A}_k
=
\begin{cases}
\Delta\mathbf{A}^{\mathrm{init}}_k \circ \tilde{\mathbf{A}}_k,
& a_k=1, \\
\tilde{\mathbf{A}}_k,
& a_k=0 .
\end{cases}
\end{equation}

\begin{algorithm}[!t]
\DontPrintSemicolon
\caption{ScalLoc Pipeline}
\label{alg:scalloc}

\SetKwInOut{Input}{Input}
\SetKwInOut{Output}{Output}

\Input{
Thermal UAV sequence $\mathcal{I}$; geo-referenced satellite map $\mathcal{M}$;
fixed intrinsic matrix $\mathbf{K}$; initial geographic state.
}
\Output{
Global trajectory $\mathcal{T}=\{(\mathbf{g}_t,\psi_t)\}_{t=1}^{N}$.
}

$\{\mathcal{W}_k\}_{k=1}^{K}\leftarrow\operatorname{BuildWindows}(\mathcal{I})$\;
$\tilde{\mathbf{A}}_1\leftarrow\operatorname{InitializePrior}(\text{initial state})$\;

\For{$k=1$ \KwTo $K$}{
    $\mathcal{P}_k,\{\mathbf{R}_{k,i},\mathbf{t}_{k,i}\},\{\mathbf{c}_{k,i}\}
    \leftarrow \operatorname{VGGTTrackRefine}(\mathcal{W}_k,\mathbf{K})$\;

    \If{$k=1$}{
        $m\leftarrow\operatorname{InferMotionType}(\{\mathbf{c}_{k,i}\})$\;
    }

    \eIf{$m=\mathrm{orbit}$}{
        $\Delta\mathbf{A}^{\mathrm{init}}_k
        \leftarrow
        \operatorname{BEVAnchor}(\mathcal{P}_k,\{\mathbf{c}_{k,i}\},
        \tilde{\mathbf{A}}_k,\mathcal{M})$\;
        $b_k\leftarrow\mathrm{BEV}$\;
    }{
        $\Delta\mathbf{A}^{\mathrm{init}}_k
        \leftarrow
        \operatorname{DirectAnchor}(\mathcal{W}_k,\mathcal{P}_k,\{\mathbf{c}_{k,i}\},
        \tilde{\mathbf{A}}_k,\mathcal{M})$\;
        $b_k\leftarrow\mathrm{dir}$\;
    }

    $a_k\leftarrow\operatorname{ValidateQuality}(\Delta\mathbf{A}^{\mathrm{init}}_k)$\;

    \If{$b_k=\mathrm{dir}$ and $a_k=0$}{
        $\Delta\mathbf{A}^{\mathrm{init}}_k,a_k,b_k
        \leftarrow
        \operatorname{RescueAnchor}(\mathcal{W}_k,\mathcal{P}_k,\{\mathbf{c}_{k,i}\},
        \tilde{\mathbf{A}}_k,\mathcal{M})$\;
    }

    \eIf{$a_k=1$}{
        $\mathbf{A}_k\leftarrow
        \Delta\mathbf{A}^{\mathrm{init}}_k\circ\tilde{\mathbf{A}}_k$\;
    }{
        $\mathbf{A}_k\leftarrow\tilde{\mathbf{A}}_k$\;
    }

    $\{(\mathbf{g}_{\tau_{k,i}},\psi_{\tau_{k,i}})\}_{i=1}^{L}
    \leftarrow
    \operatorname{RecoverWindowTrajectory}(\mathbf{A}_k,\{\mathbf{c}_{k,i}\},b_k)$\;

    \If{$k<K$}{
        $\tilde{\mathbf{A}}_{k+1}
        \leftarrow
        \operatorname{PropagatePrior}(\mathbf{A}_k,\mathcal{W}_{k+1})$\;
    }
}

\textbf{return} $\mathcal{T}=\{(\mathbf{g}_t,\psi_t)\}_{t=1}^{N}$\;

\end{algorithm}

After $\mathbf{A}_k$ is determined, it is applied to the local camera trajectory
to recover frame-level UTM positions. Since $\mathbf{A}_k$ is a planar
local-to-map transformation, each 3D camera center $\mathbf{c}_{k,i}$ is first
converted to the same local 2D source coordinate $\mathbf{p}_{k,i}$ used by the retained anchoring
branch:

\begin{equation}
\mathbf{p}_{k,i}
=
\begin{cases}
\Pi^{\mathrm{BEV}}_k(\mathbf{c}_{k,i}), & \text{BEV branch},\\
\Pi^{\mathrm{dir}}_k(\mathbf{c}_{k,i}), & \text{direct branch}.
\end{cases}
\end{equation}

The frame-level UTM position is then obtained as:

\begin{equation}
\mathbf{g}_{k,i}
=
\mathbf{A}_k
\left(
\mathbf{p}_{k,i}
\right).
\end{equation}

The heading of each frame is recovered from its local forward direction. Let
$\mathbf{f}_{k,i}\in\mathbb{R}^{2}$ denote the nonzero local forward direction
of frame $i$, and let $\mathbf{S}_k\in\mathbb{R}^{2\times2}$ be the linear
component of $\mathbf{A}_k$. We denote the transformed UTM direction vector by
$\mathbf{v}_{k,i}=\mathbf{S}_k\mathbf{f}_{k,i}$, whose east and north components
are $v^E_{k,i}$ and $v^N_{k,i}$, respectively. The global heading
$\psi_{k,i}$ is:
\begin{equation}
\psi_{k,i}
=
\operatorname{wrap}
\left(
\operatorname{atan2}
\left(
v^E_{k,i},
v^N_{k,i}
\right)
\right),
\end{equation}
where $\operatorname{atan2}(y,x)$ denotes the two-argument arctangent that
returns the signed angle of a 2D vector while preserving its quadrant, and
$\operatorname{wrap}(\cdot)$ maps angles to $(-\pi,\pi]$.

The final state of $\mathcal{W}_k$ is used to construct the pre-anchor prior
for the next window. For the first window, $\tilde{\mathbf{A}}_1$ is initialized
from the first-frame geographic position, heading, and relative height. For
subsequent windows, adjacent windows share the boundary frame
$\tau_{k,L}=\tau_{k+1,1}$. Therefore, the corrected position and heading of the
last frame in $\mathcal{W}_k$ provide the global reference for the first frame
of $\mathcal{W}_{k+1}$.

The propagated prior $\tilde{\mathbf{A}}_{k+1}$ is constructed from
$\mathbf{A}_k$ by inheriting the current scale and enforcing position and
heading continuity at the shared boundary frame:

\begin{equation}
\tilde{\mathbf{A}}_{k+1}
:
\left(
\mathbf{p}_{k+1,1},
\mathbf{f}_{k+1,1}
\right)
\Leftarrow
\left(
\mathbf{g}_{k,L},
\psi_{k,L}
\right),
\end{equation}
where $\mathbf{p}_{k+1,1}$ and $\mathbf{f}_{k+1,1}$ are the local 2D camera
coordinate and local forward direction of the first frame in
$\mathcal{W}_{k+1}$, and $\mathbf{g}_{k,L}$ and $\psi_{k,L}$ are the corrected
UTM position and heading of the shared boundary frame from $\mathcal{W}_k$.
This validated window-to-window propagation provides the pre-anchor global pose
prediction for the next satellite anchoring stage.

Frame estimates are indexed by their global frame indices. Specifically, for
$t=\tau_{k,i}$, we set $\mathbf{g}_t=\mathbf{g}_{k,i}$ and
$\psi_t=\psi_{k,i}$. The final trajectory is:
\begin{equation}
\mathcal{T}
=
\left\{
(\mathbf{g}_t,\psi_t)
\right\}_{t=1}^{N}.
\end{equation}

The complete workflow of ScalLoc is shown in Algorithm~\ref{alg:scalloc}.

\section{Experimental Setup}
\label{sec:Experimental Setup}

\begin{figure*}[!t]
    \centering
    \includegraphics[width=\textwidth]{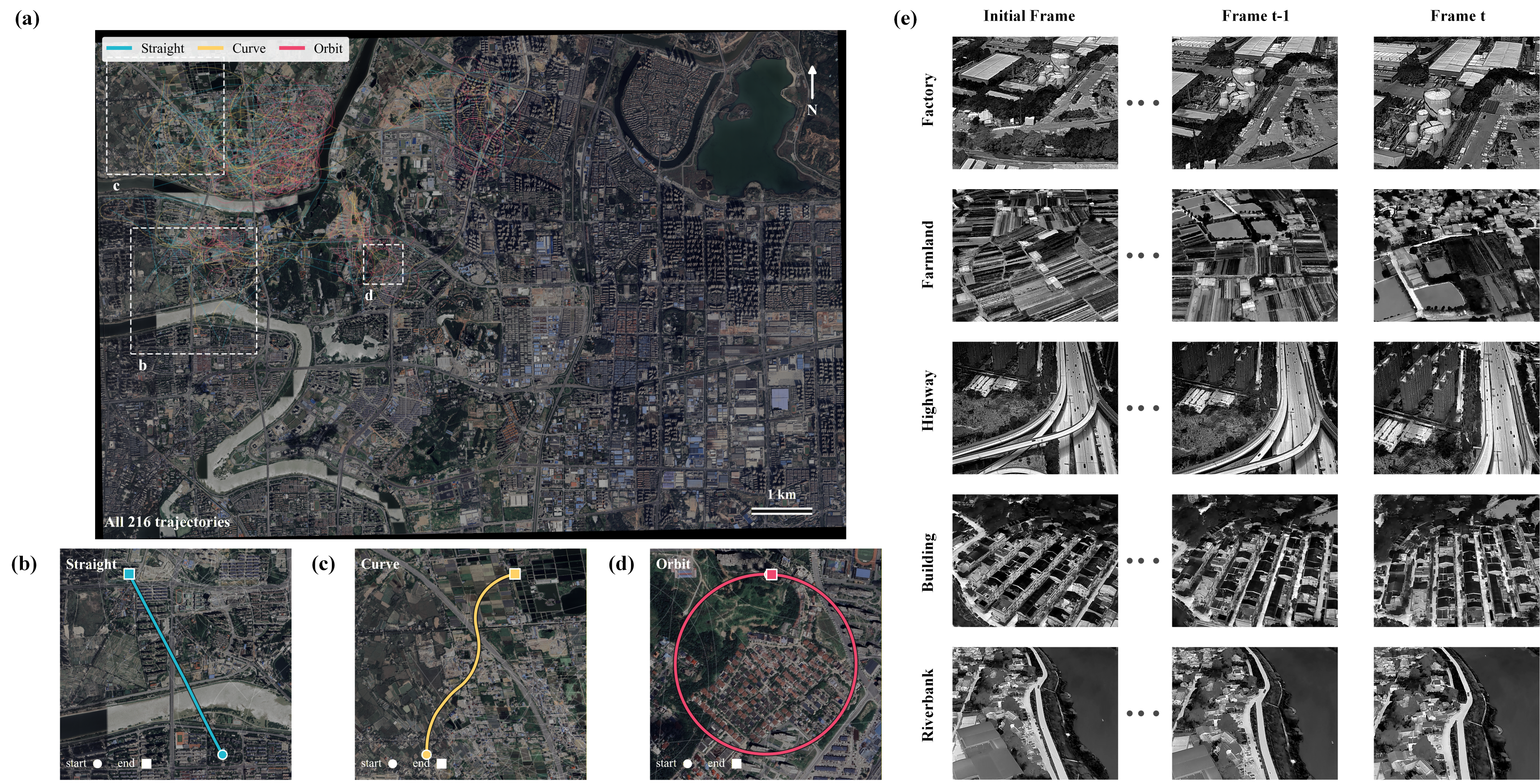}
    \caption{Overview of the Thermal-Oblique dataset. (a) Spatial distribution of 216 trajectories. (b)--(d) Representative straight, curve, and orbit trajectories. (e) Example thermal sequences from different scene categories.}
    \label{fig:thermal_oblique}
\end{figure*}

\subsection{Datasets}
\label{sec:datasets}

We evaluate continuous oblique thermal geo-localization on the proposed \textbf{Thermal-Oblique} dataset. As shown in Fig.~\ref{fig:thermal_oblique}, it contains real aerial thermal videos captured over rural and urban scenes, including factory, farmland, highway, building, and riverbank areas. The dataset was collected in Changsha, Hunan, China using a DJI Matrice 4T platform\footnote{\url{https://www.dji.com/support/product/}} equipped with a thermal camera. During data collection, we record continuous thermal videos and then sample one frame per second from each video clip. The sampled thermal frames have a resolution of $1280\times1040$ and are synchronized with onboard telemetry. In total, Thermal-Oblique contains 216 video clips and 28,112 sampled thermal frames.

Thermal-Oblique is designed as a controlled benchmark covering variations in scene type, acquisition time, motion pattern, relative flight height, and camera pitch. It includes rural and urban scenes under day and night conditions, three motion patterns, i.e., straight, curve, and orbit, three relative heights above the take-off ground plane, 150, 200, and 250~m, and three downward pitch angles measured from the horizontal plane, $30^\circ$, $35^\circ$, and $40^\circ$. For straight and curve trajectories, each UAV video is recorded for approximately 2 minutes. For orbit trajectories, each video covers one complete circular flight, with a duration of approximately 2--4 minutes. Each combination is represented by two clips, forming a balanced evaluation design.

Each sampled frame is associated with detailed onboard pose information, including geographic position, relative take-off altitude, and gimbal attitude. These frame-level annotations support research on sequential oblique thermal UAV geo-localization and enable evaluation of trajectory recovery, heading estimation, and long-horizon localization stability. Geo-referenced satellite maps and DSM data are obtained from Thermal-UAV~\cite{zhang2026scc}. The satellite maps are used as the global 2D reference for all evaluated methods, while the DSM is used only for DSM-dependent baselines and is not used by ScalLoc.

\begin{figure*}[!t]
    \centering
    \includegraphics[width=\textwidth]{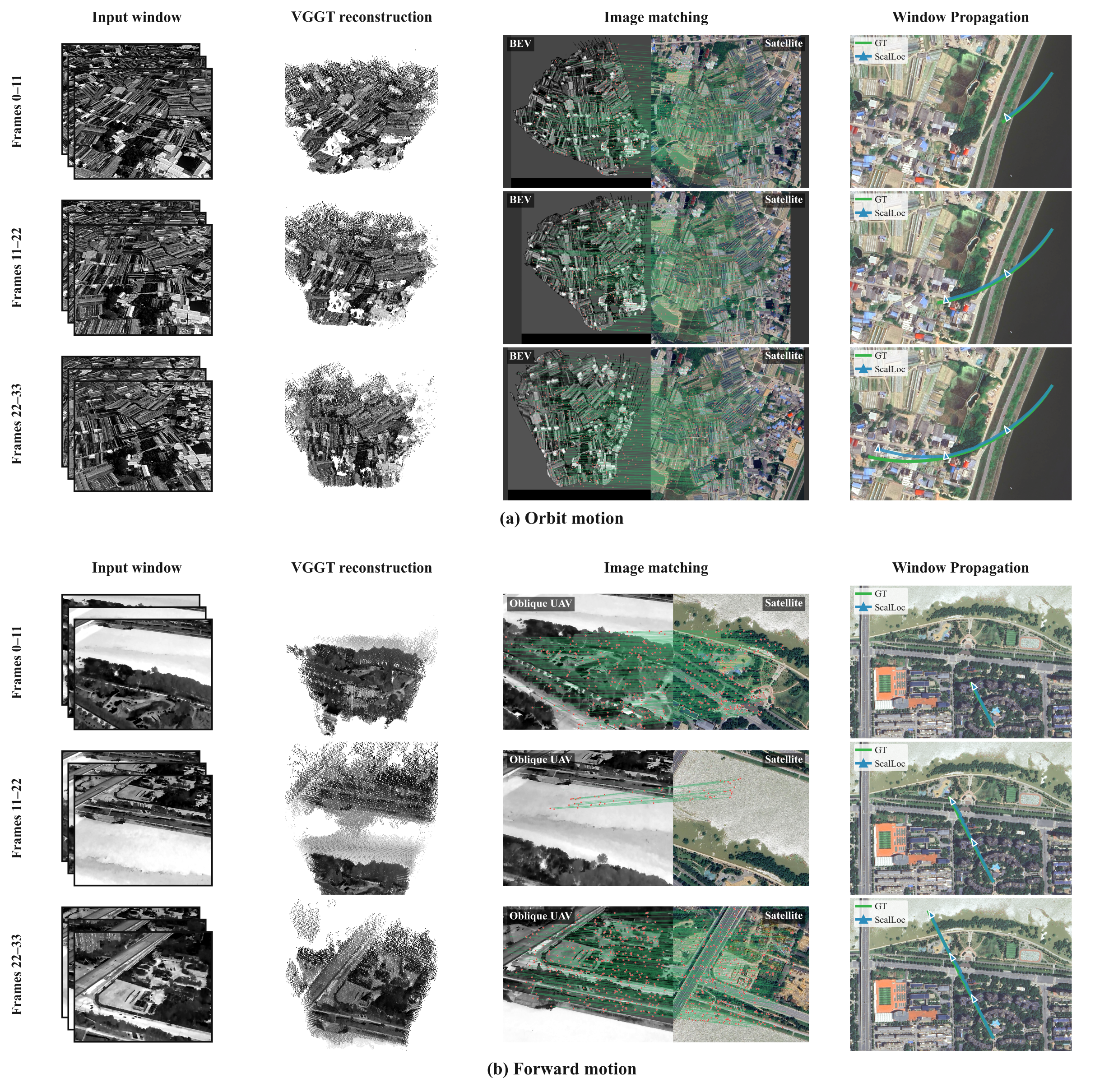}
    \caption{Qualitative operation examples of ScalLoc. (a) Orbit motion uses BEV-to-satellite matching by converting local VGGT reconstructions into BEV representations. (b) Forward motion uses direct oblique image-to-satellite matching with selected thermal frames and satellite tiles. Both branches propagate validated anchor states to recover continuous trajectories.}
    \label{fig:operation_examples}
\end{figure*}

\begin{figure}[!t]
    \centering
    \includegraphics[width=\linewidth]{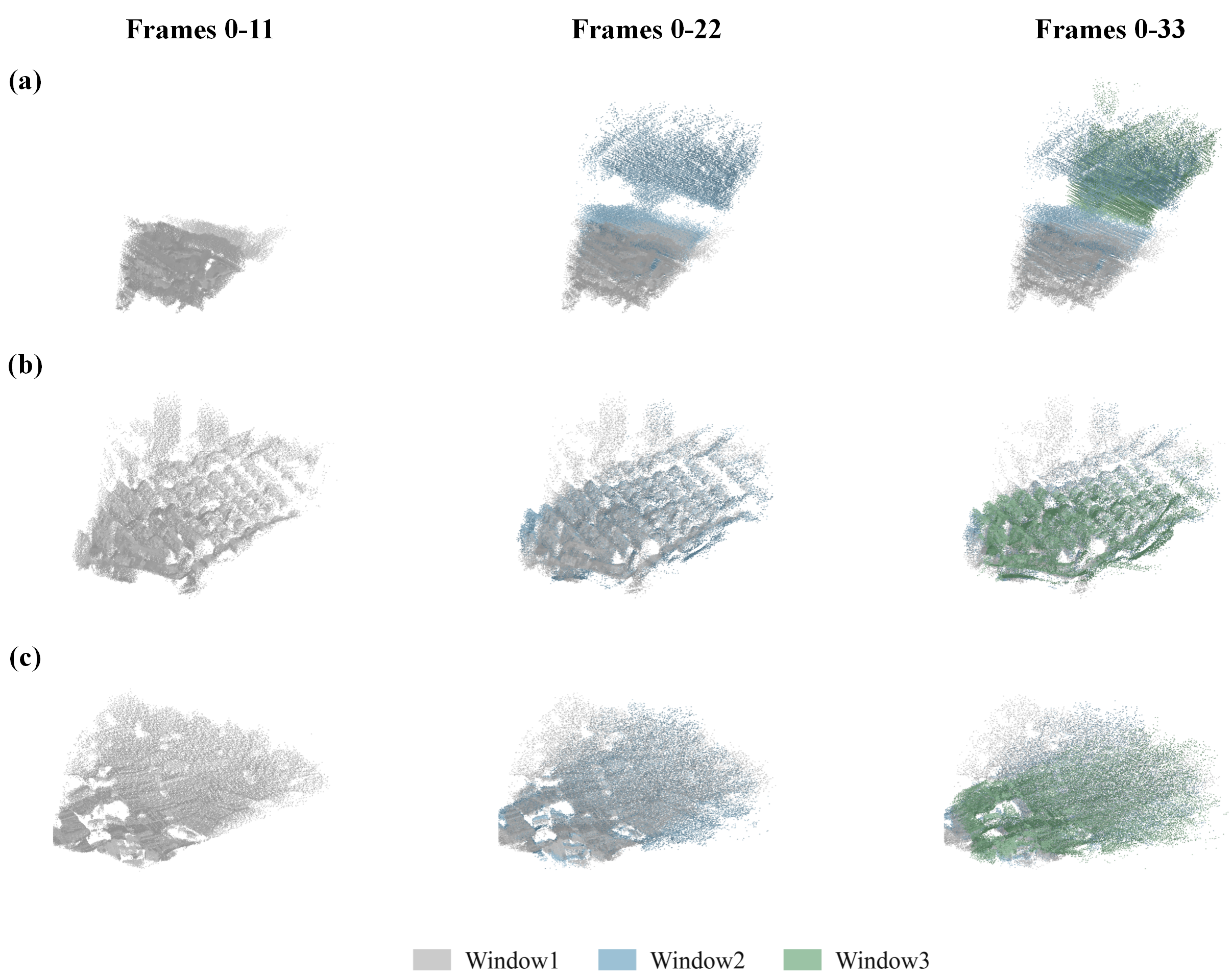}
    \caption{
    Globally anchored point-cloud merging in a shared coordinate frame.
    Rows show (a) straight, (b) curve, and (c) orbit trajectories, while columns show
    progressive window merging. Different colors denote different windows.
    }
    \label{fig:pointcloud_merging}
\end{figure}

\subsection{Evaluation Metrics}
\label{sec:evaluation_metrics}

We evaluate localization performance at the frame level using position and heading errors. The position error is defined as the Euclidean distance between the predicted UTM position and the reference UTM position, measured in meters~\cite{xiao2024sthn}. Based on this error, we report \textbf{Pos. Acc@5/10/20}, i.e., the percentage of frames whose position errors are within 5, 10, and 20~m, respectively. The heading error is the wrapped absolute angular difference between the predicted and reference headings, measured in degrees. We report \textbf{Yaw Acc@3/5/10}, i.e., the percentage of frames whose heading errors are within $3^\circ$, $5^\circ$, and $10^\circ$, respectively.

For each trajectory pattern, Pos. Acc@5/10/20 and Yaw Acc@3/5/10 are computed over all frames belonging to that pattern. When multiple trajectories are involved, we aggregate results at the frame level, so trajectories with more frames contribute proportionally more to the reported statistics. We also report the \textbf{Mean Error (ME)}, \textbf{Standard Deviation (SD)}, and \textbf{Median Error (MedE)} of frame-level position errors, all measured in meters~\cite{dhaouadi2026orthotrack}. ME reflects the average localization error, SD measures error dispersion, and MedE describes the typical error while being less sensitive to long-tailed failures.

We further report the average online processing time per frame, denoted as \textbf{Time Per Frame}. It is measured in seconds per frame and excludes data loading and offline reference-map preparation, reflecting only the online localization stage.

\subsection{Baseline Methods}
\label{sec:baseline_methods}

We compare ScalLoc with external baselines and controlled variants, grouped by how global geographic information is introduced. \textbf{DSM-assisted geo-localization methods} use satellite imagery together with map-side elevation. This group includes \textbf{SCC-Loc}~\cite{zhang2026scc}, \textbf{OrthoLoC}~\cite{dhaouadi2025ortholoc}, and \textbf{OrthoTrack}~\cite{dhaouadi2026orthotrack}, which use DSM/elevation references that are not used by ScalLoc.

\textbf{Scale-aligned sequence-only methods} estimate relative camera motion from thermal sequences and recover global trajectories using the initial geographic state and a fixed scale, without subsequent satellite-map correction. This group includes \textbf{DROID-SLAM}~\cite{teed2021droid}, \textbf{ORB-SLAM3}~\cite{campos2021orb}, and \textbf{VGGT-SLAM}~\cite{wang2025vggt}. They evaluate whether visual odometry or sequence reconstruction alone can maintain accurate long-horizon global localization.

We also evaluate three \textbf{satellite-corrected ScalLoc variants}. \textbf{BEV-only} applies BEV-to-satellite alignment to all trajectory types, while \textbf{Direct-only} applies direct multi-frame image-to-satellite alignment to all trajectory types. \textbf{Motion-adaptive} denotes the complete ScalLoc method, which selects the anchoring branch according to the trajectory pattern and uses quality validation, rescue, and window-to-window propagation.

\subsection{Implementation Details}
\label{sec:implementation_details}

\textbf{Implementation Platform.} ScalLoc is implemented in PyTorch and evaluated on a workstation equipped with 25 vCPUs of an Intel Xeon Gold 6459C processor and one NVIDIA RTX 5090 GPU with 32~GB memory.

\textbf{Parameter Settings.} Unless otherwise specified, each temporal window contains $L=12$ frames. In the direct multi-frame anchoring branch, we select three keyframes from each window for image-to-satellite matching, and the satellite tile size is set to $1024\times1024$ pixels. These parameters are kept fixed for all experiments.

\textbf{Evaluation Protocol.} All methods are evaluated under the same initial-state setting, where the first frame of each trajectory is given with a known geographic pose. Subsequent poses are estimated by either satellite-map anchoring or trajectory propagation from this initial state. For SCC-Loc, we follow its single-frame PnP-based localization protocol. If localization fails on the default satellite crop, we perform retrieval within a local satellite tile centered at the predicted pose, followed by matching and PnP estimation. OrthoLoC is evaluated frame by frame using its DSM-assisted 2D--3D correspondence and PnP-based pose estimation protocol. For sequence-only baselines, including DROID-SLAM, ORB-SLAM3, and VGGT-SLAM, the trajectory is first estimated in a local coordinate system. We compute the metric scale as the ratio of the real relative height to the corresponding local relative height, and propagate the scaled relative poses from the known first-frame pose to obtain global positions. No subsequent satellite-map correction is applied to these sequence-only baselines.

\section{Results}
\label{sec:Results}

\begin{table*}[!t]
\centering
\caption{
Quantitative comparison under different trajectory patterns on Thermal-Oblique.
The best and second-best results are highlighted in \textbf{bold} and
\underline{underlined}, respectively.
}
\label{tab:baseline_comparison_by_motion}

\renewcommand{\arraystretch}{1.45}
\setlength{\tabcolsep}{3.2pt}

\resizebox{\textwidth}{!}{%
\begin{tabular}{@{}lccccccccc@{}}
\toprule

\multirow{2}{*}{\makecell[l]{\textbf{Method}}}
& \multicolumn{2}{c}{\textbf{Straight}}
& \multicolumn{2}{c}{\textbf{Curve}}
& \multicolumn{2}{c}{\textbf{Orbit}}
& \multirow{2}{*}{\makecell[c]{\textbf{All Frames}\\\textbf{ME $\pm$ SD}}}
& \multirow{2}{*}{\makecell[c]{\textbf{All Frames}\\\textbf{MedE}}}
& \multirow{2}{*}{\makecell[c]{\textbf{Time}\\\textbf{Per Frame}}}
\\

\cmidrule(lr){2-3}
\cmidrule(lr){4-5}
\cmidrule(lr){6-7}

&
\makecell[c]{\textbf{Pos.}\\\textbf{Acc@5/10/20}}
& \makecell[c]{\textbf{Yaw}\\\textbf{Acc@3/5/10}}
& \makecell[c]{\textbf{Pos.}\\\textbf{Acc@5/10/20}}
& \makecell[c]{\textbf{Yaw}\\\textbf{Acc@3/5/10}}
& \makecell[c]{\textbf{Pos.}\\\textbf{Acc@5/10/20}}
& \makecell[c]{\textbf{Yaw}\\\textbf{Acc@3/5/10}}
& & & \\
\midrule

\rowcolor{gray!12}
\multicolumn{10}{@{}l}{\textbf{A. DSM-assisted Geo-localization Methods}} \\

\hspace{0.6em}SCC-Loc~\cite{zhang2026scc}
& 16.53 / 31.50 / 43.70
& 28.55 / 44.27 / 55.47
& 19.23 / 36.86 / 51.08
& 29.03 / 49.30 / 63.46
& \underline{27.66} / \underline{51.06} / \underline{65.64}
& 32.37 / 58.17 / 74.87
& $566.97 \pm 1317.47$
& 15.43
& 1.97 \\

\hspace{0.6em}OrthoLoC~\cite{dhaouadi2025ortholoc}
& 23.16 / 40.15 / 49.35
& 29.40 / 47.26 / 57.32
& \textbf{26.19} / \textbf{44.50} / 55.37
& 27.94 / 49.35 / 62.97
& \textbf{34.91} / \textbf{60.85} / \textbf{74.86}
& 37.25 / 64.60 / 81.87
& $522.81 \pm 1237.20$
& \textbf{10.36}
& 1.36 \\

\hspace{0.6em}OrthoTrack~\cite{dhaouadi2026orthotrack}
& \textbf{31.12} / 46.92 / 55.62
& 45.32 / 67.85 / 87.18
& \underline{21.47} / 33.15 / 39.00
& 24.08 / 38.27 / 50.38
& 21.13 / 41.92 / 58.33
& 34.16 / 55.43 / 75.09
& $234.28 \pm 367.51$
& 17.85
& 0.55 \\

\midrule

\rowcolor{gray!12}
\multicolumn{10}{@{}l}{\textbf{B. Scale-aligned Sequence-only Methods}} \\

\hspace{0.6em}DROID-SLAM~\cite{teed2021droid}
& 4.85 / 8.62 / 16.64
& \textbf{99.27} / \textbf{100.00} / \textbf{100.00}
& 3.52 / 6.14 / 9.91
& \textbf{55.36} / \textbf{78.15} / \textbf{97.35}
& 9.68 / 18.57 / 35.33
& \textbf{94.61} / \textbf{99.99} / \textbf{100.00}
& $72.44 \pm \mathbf{66.76}$
& 48.72
& \underline{0.10} \\

\hspace{0.6em}ORB-SLAM3~\cite{campos2021orb}
& 3.59 / 6.06 / 10.58
& \underline{93.47} / 95.71 / 98.59
& 2.75 / 4.70 / 7.72
& 15.12 / 20.03 / 30.31
& 3.46 / 6.21 / 12.72
& 31.99 / 34.34 / 36.85
& $429.08 \pm 416.71$
& 323.07
& \textbf{0.05} \\

\hspace{0.6em}VGGT-SLAM~\cite{wang2025vggt}
& 7.48 / 12.80 / 20.70
& 90.89 / \underline{96.14} / 97.83
& 6.57 / 11.32 / 17.36
& 40.60 / 59.74 / 87.41
& 7.47 / 12.32 / 24.85
& \underline{53.63} / \underline{71.56} / \underline{92.69}
& $110.06 \pm 157.49$
& 58.31
& 0.46 \\

\midrule

\rowcolor{gray!12}
\multicolumn{10}{@{}l}{\textbf{C. Satellite-corrected ScalLoc Variants}} \\

\hspace{0.6em}BEV-only
& 8.85 / 20.52 / 40.80
& 56.30 / 81.60 / 99.31
& 7.44 / 16.81 / 37.45
& 43.36 / 63.49 / 90.48
& 10.75 / 25.65 / 51.89
& 42.37 / 62.58 / 85.46
& $51.61 \pm 100.09$
& 23.33
& 0.51 \\

\hspace{0.6em}Direct-only
& 28.84 / \underline{55.56} / \underline{80.63}
& 49.84 / 87.71 / 99.54
& 16.66 / \underline{38.81} / \textbf{74.96}
& \underline{44.95} / 75.38 / \underline{96.15}
& 8.06 / 14.84 / 27.55
& 26.90 / 40.94 / 61.31
& $\underline{35.53} \pm \underline{66.86}$
& 15.32
& 0.55 \\

\hspace{0.6em}Motion-adaptive (Ours)
& \underline{28.87} / \textbf{56.17} / \textbf{81.86}
& 50.06 / 87.63 / \underline{99.57}
& 16.86 / 38.58 / \underline{74.81}
& 44.91 / \underline{75.40} / 96.11
& 9.19 / 25.93 / 53.51
& 42.01 / 62.99 / 82.67
& $\mathbf{26.97} \pm 71.67$
& \underline{12.66}
& 0.56 \\

\bottomrule
\end{tabular}%
}
\end{table*}

\subsection{Illustrative Operation}
\label{sec:illustrative_operation}

Fig.~\ref{fig:operation_examples} illustrates representative operating cases of ScalLoc under orbit-like and forward motion. In both cases, VGGT produces locally coherent 3D reconstructions from oblique thermal image windows, although it is trained mainly on RGB data~\cite{wang2025vggt}. This indicates that the learned geometric representation remains effective for thermal imagery and provides usable local geometry for subsequent satellite anchoring.

For orbit-like motion, as shown in Fig.~\ref{fig:operation_examples}(a), ScalLoc converts the reconstructed point map into a BEV orthophoto and matches it with the satellite map. The BEV representation reduces the viewpoint gap caused by rotational motion and provides stable satellite correction, allowing the recovered trajectory to remain close to the reference loop. For forward motion, as shown in Fig.~\ref{fig:operation_examples}(b), ScalLoc performs direct oblique image-to-satellite matching. The middle example contains many unreliable matches on the water region, leading to an invalid satellite correction. This correction is rejected by quality validation, so the propagated prior is retained and the trajectory is not contaminated by the failed match.

Fig.~\ref{fig:pointcloud_merging} further visualizes how local window reconstructions are aligned through satellite anchors. Each color denotes the point cloud reconstructed from a different temporal window, and the columns show the progressive merging process after additional windows are anchored. The straight trajectory shows point clouds extending forward along the flight direction, the curve trajectory shows rightward overlap as the UAV turns, and the orbit trajectory shows stacked point clouds around the same local area. These patterns indicate that the accepted anchors align local windows into a shared coordinate system while preserving different motion structures, supporting continuous global trajectory recovery.

\begin{figure}[!t]
    \centering
    \includegraphics[width=\linewidth]{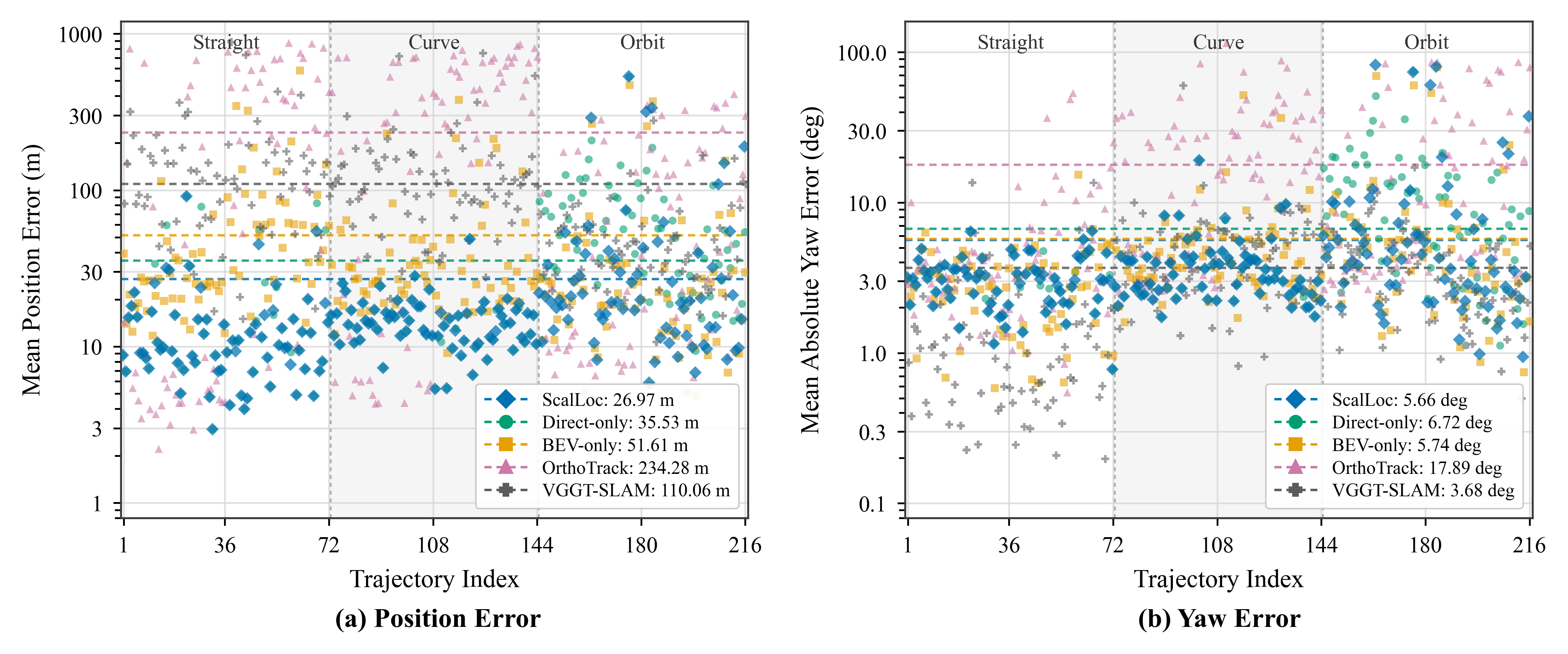}
    \caption{
    Trajectory-wise position and yaw errors on Thermal-Oblique.
    Dashed lines denote frame-weighted mean errors.
    }
    \label{fig:trajectory_error_scatter}
\end{figure}

\begin{figure*}[!t]
    \centering
    \includegraphics[width=\textwidth]{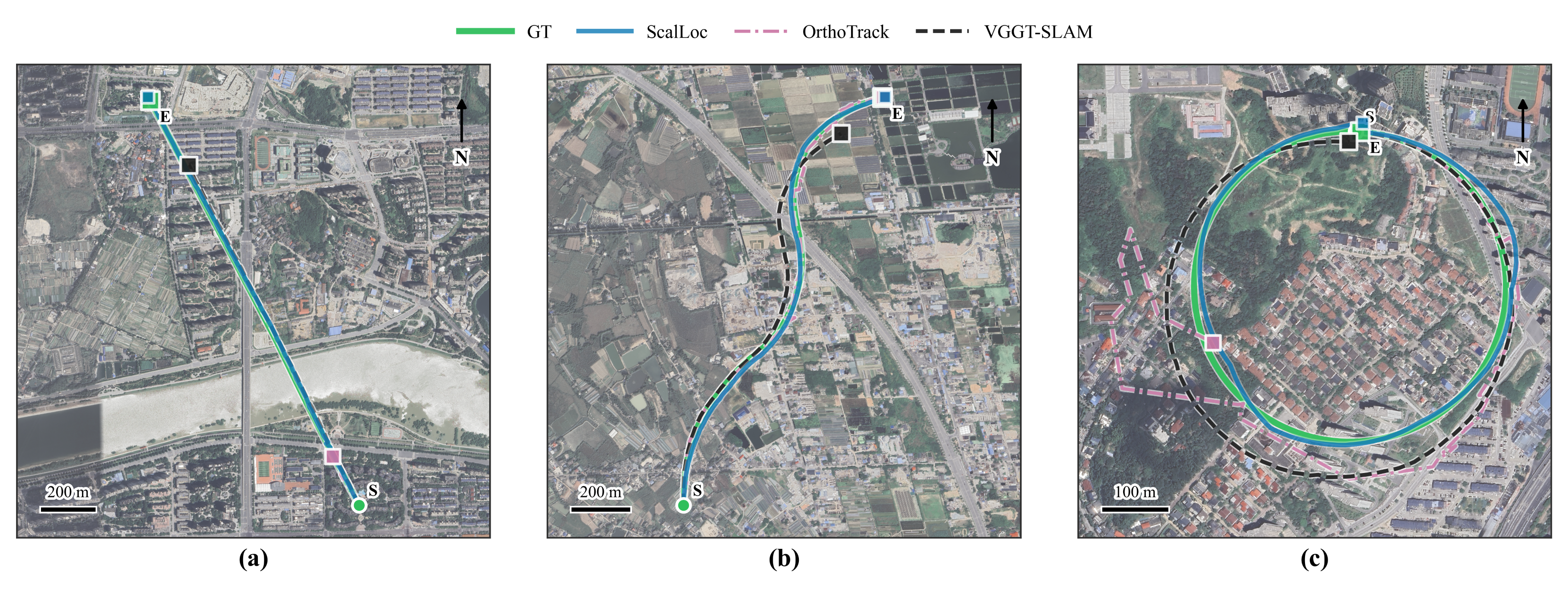}
    \caption{
    Qualitative trajectory comparison on Thermal-Oblique. 
    (a) Straight, (b) curve, and (c) orbit trajectories. The reference trajectory is
    shown in green, ScalLoc in blue, OrthoTrack in pink, and VGGT-SLAM in black.
    ScalLoc stays close to the reference trajectory across different motion patterns.
    }
    \label{fig:compare_with_other}
\end{figure*}

\subsection{Comparison with Existing Methods}
\label{sec:comparison_existing_methods}

Table~\ref{tab:baseline_comparison_by_motion} reports the quantitative comparison under straight, curve, and orbit trajectories. DSM-assisted methods obtain competitive frame-level accuracies in several cases, especially OrthoLoC on orbit trajectories and OrthoTrack on straight trajectories. OrthoLoC also achieves the lowest all-frame MedE of 10.36~m, showing that DSM-based 2D--3D geometry can provide accurate estimates when reliable correspondences and high-quality map-side geometry are available. However, SCC-Loc and OrthoLoC exhibit very large ME and SD despite their low median errors. This gap indicates occasional severe failures, which can arise from the strong dependence on accurate DSM geometry, satellite--DSM registration, and reliable cross-modal 2D--3D correspondences. Errors in these components can be directly transferred to the absolute pose under oblique thermal views. In contrast, sequence-only methods preserve orientation well, especially DROID-SLAM, but accumulate position errors without recurring satellite-map correction.

The ScalLoc variants reveal the importance of satellite correction and motion-adaptive branch selection. BEV-only provides stable alignment for orbit-like motion but loses oblique-view cues for forward-moving trajectories. Direct-only performs better on straight and curve trajectories but degrades on orbit trajectories. Motion-adaptive ScalLoc combines the two anchoring strategies and validates corrections before propagation. Although its MedE of 12.66~m is slightly higher than that of OrthoLoC, it achieves the lowest all-frame ME of 26.97~m and substantially reduces large-error failures. Fig.~\ref{fig:trajectory_error_scatter} further shows that sequence-only methods exhibit large position drift, while DSM-assisted methods and single-branch variants contain large-error outliers. ScalLoc suppresses these failures over most trajectories through repeated satellite-map anchoring and quality validation.


Fig.~\ref{fig:compare_with_other} presents representative trajectory comparisons under straight, curve, and orbit motions. In (a) and (c), OrthoTrack suffers from erroneous sequential pose estimates, and the error propagates to subsequent frames, causing large trajectory deviations. In (b) and (c), VGGT-SLAM preserves a trajectory shape similar to the reference after fixed-scale alignment, but still exhibits global drift without satellite-map correction after initialization. In contrast, ScalLoc repeatedly anchors local windows to the satellite map and validates corrections before propagation, keeping the recovered trajectory close to the reference across different motion patterns.

\subsection{Ablation Study}
\label{sec:ablation}

To analyze the contribution of the main components in ScalLoc, we conduct ablation studies on the full Thermal-Oblique dataset. Table~\ref{tab:ablation_vggt_ba_anchor} compares five variants: \textit{VGGT}, \textit{VGGT+Refine}, \textit{VGGT+Anchor}, \textit{VGGT+Refine+Anchor (Without Validation)}, and the complete \textit{VGGT+Refine+Anchor} pipeline.

The results show that satellite anchoring is the key factor for removing global drift. Pure VGGT and \textit{VGGT+Refine} recover only local geometry, so their trajectories remain in a local frame and accumulate large global errors. Adding satellite anchoring greatly reduces ME from 112.69~m to 36.15~m, because repeated local-to-map corrections provide absolute geographic constraints. Track-based pose refinement alone brings a modest gain, but when combined with anchoring, it further reduces ME to 26.97~m. This indicates that pose refinement mainly improves local camera geometry, making subsequent satellite matching and residual fitting more reliable.

Anchor validation is critical for preventing error propagation. Without validation, all estimated anchor corrections are propagated to later windows, including unreliable matches caused by weak texture, repetitive structures, or incorrect satellite correspondences. As a result, ME increases to 76.97~m and MedE increases to 45.66~m. The complete pipeline rejects low-quality corrections and retains the propagated prior when anchoring is unreliable, which prevents local matching failures from contaminating long-horizon trajectories.

Table~\ref{tab:runtime_breakdown} reports the runtime breakdown on representative trajectories. VGGT reconstruction and satellite anchoring dominate the per-window runtime, while track-based pose refinement adds little overhead. The orbit case has lower anchor time because the BEV branch uses a compact orthophoto matching process, whereas straight and curve cases rely more on direct multi-frame oblique image-to-satellite matching. Overall, ScalLoc runs at about 0.44--0.57 seconds per frame under the evaluated settings.

\begin{table}[!t]
\centering
\caption{Ablation results of VGGT reconstruction, track-based pose refinement, and satellite anchoring on Thermal-Oblique.}
\label{tab:ablation_vggt_ba_anchor}
\renewcommand{\arraystretch}{1.3}
\setlength{\tabcolsep}{3.5pt}
\resizebox{\columnwidth}{!}{%
\begin{tabular}{lcccc}
\toprule
\textbf{Method}
& \makecell{\textbf{Pos.}\\\textbf{Acc@5/10/20}}
& \makecell{\textbf{Yaw}\\\textbf{Acc@3/5/10}}
& \textbf{ME $\pm$ SD}
& \textbf{MedE} \\
\midrule
VGGT
& 2.78 / 5.58 / 12.46
& 53.14 / 64.24 / 77.50
& $112.69 \pm 110.09$
& 78.98 \\

VGGT + Refine
& 4.70 / 8.25 / 16.28
& \textbf{56.34} / 67.57 / 80.31
& $103.82 \pm 107.68$
& 72.36 \\

VGGT + Anchor
& 12.74 / 31.26 / 61.21
& 42.51 / 71.02 / 88.96
& $36.15 \pm 82.08$
& 15.54 \\

\makecell[l]{VGGT + Refine + Anchor \\ (Without Validation)}
& 3.81 / 15.46 / 30.18
& 60.32 / 70.32 / 83.25
& $76.97 \pm 213.67$
& 45.66 \\

\makecell[l]{VGGT + Refine + Anchor (Ours)}
& \textbf{17.81 / 39.46 / 69.18}
& 45.46 / \textbf{74.68} / \textbf{92.25}
& $\mathbf{26.97 \pm 71.67}$
& \textbf{12.66} \\

\bottomrule
\end{tabular}%
}
\end{table}

\begin{table}[!t]
\centering
\caption{
Runtime breakdown of ScalLoc on representative trajectories.
All times are reported in seconds.
}
\label{tab:runtime_breakdown}
\renewcommand{\arraystretch}{1.15}
\setlength{\tabcolsep}{7.5pt}
\resizebox{\linewidth}{!}{%
\begin{tabular}{lccccc}
\toprule
\textbf{Trajectory}
& \makecell{\textbf{VGGT}\\\textbf{Time}}
& \makecell{\textbf{Refine}\\\textbf{Time}}
& \makecell{\textbf{Anchor}\\\textbf{Time}}
& \makecell{\textbf{Window}\\\textbf{Time}}
& \makecell{\textbf{Avg.}\\\textbf{Frame Time}} \\
\midrule
Straight
& 2.87
& 0.08
& 2.88
& 6.73
& 0.57 \\
Curve
& 2.65
& 0.06
& 2.79
& 6.32
& 0.56 \\
Orbit
& 2.66
& 0.06
& 1.74
& 5.22
& 0.44 \\
\bottomrule
\end{tabular}%
}
\end{table}

\begin{figure*}[!t]
    \centering
    \includegraphics[width=0.9\textwidth]{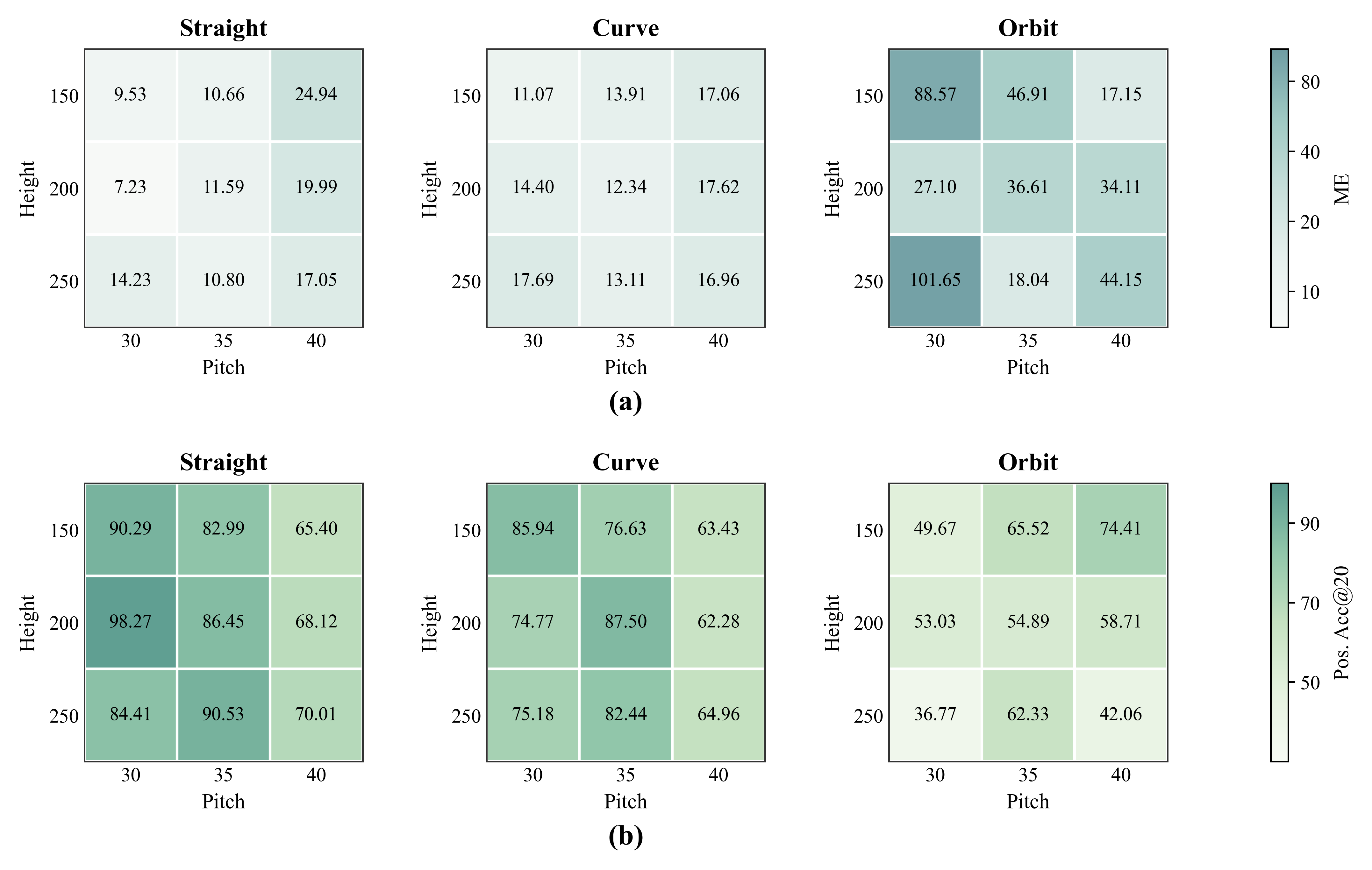}
    \caption{
    Performance of ScalLoc across flight heights and camera pitch angles.
    The first row shows ME in meters, and the second row shows Pos. Acc@20 for
    straight, curve, and orbit trajectories.
    }   
    \label{fig:height_pitch_heatmap}
\end{figure*}

\subsection{Data Factor Analysis}
\label{sec:data_factor_analysis}

We analyze how data factors affect ScalLoc, including flight height, camera pitch, motion pattern, scene type, and lighting condition. Fig.~\ref{fig:height_pitch_heatmap} first analyzes the joint influence of flight height, camera pitch, and motion pattern. Straight and curve trajectories remain relatively stable across most height--pitch settings because forward motion preserves useful oblique-view context for direct matching. The $40^\circ$ pitch setting tends to degrade performance in several cases, indicating that overly downward views reduce forward-looking context and matching coverage. Orbit trajectories are more sensitive to height and pitch because viewpoint rotation changes the projected footprint and overlap with the satellite map more strongly. These results suggest that motion pattern is the primary factor, while height and pitch further influence localization through footprint scale, visible context, and cross-view overlap.

Table~\ref{tab:scene_lighting} then reports the results under urban/rural and day/night settings. ScalLoc achieves comparable performance in urban day and night scenes, suggesting that thermal observations provide relatively stable structural cues under illumination changes. In rural scenes, the median error remains moderate, but the larger ME and SD under nighttime conditions indicate long-tailed failures. This is likely because rural nighttime scenes contain fewer distinctive man-made structures and lower thermal contrast, which weakens satellite--thermal matching and can also reduce the stability of local reconstruction.

\begin{table}[t]
\centering
\caption{Performance across scene and lighting conditions.}
\label{tab:scene_lighting}
\renewcommand{\arraystretch}{1.3}
\setlength{\tabcolsep}{4pt}

\resizebox{\columnwidth}{!}{%
\begin{tabular}{llcccc}
\toprule

\textbf{Scene}
& \textbf{Lighting}
& \makecell{\textbf{Pos.}\\\textbf{Acc@5/10/20}}
& \makecell{\textbf{Yaw}\\\textbf{Acc@3/5/10}}
& \textbf{ME $\pm$ SD}
& \textbf{MedE} \\

\midrule

\multirow{2}{*}{Urban}
& Day
& 24.54 / 49.77 / 76.35
& 43.11 / 78.15 / 93.44
& $24.31 \pm 65.41$
& 10.07 \\

& Night
& 19.11 / 44.05 / 74.32
& 55.23 / 81.36 / 94.84
& $23.01 \pm 49.90$
& 11.26 \\

\midrule

\multirow{2}{*}{Rural}
& Day
& 12.33 / 29.56 / 63.02
& 42.09 / 72.08 / 92.61
& $19.98 \pm 17.87$
& 15.59 \\

& Night
& 15.57 / 35.06 / 63.58
& 41.88 / 67.67 / 88.33
& $40.03 \pm 113.55$
& 14.71 \\

\bottomrule
\end{tabular}%
}
\end{table}

\subsection{Parameter Analysis}
\label{sec:parameter_analysis}

Table~\ref{tab:window_keyframe_analysis_by_motion} analyzes the effects of window size and keyframe count under different motion patterns. For straight trajectories, increasing the keyframe count generally reduces ME because multiple keyframes provide broader satellite--thermal matching coverage and reduce the risk of single-frame matching failure. A moderate window size is still preferred: a very short window provides limited reconstruction context, while an overly long window reduces the frequency of satellite correction and may accumulate local pose errors before the next anchor.

For curve trajectories, the best ME appears with a smaller window. This is because curved motion involves continuous heading changes, and a larger window may make the recovered relative poses and the predicted satellite crops less accurate, causing crop shifts and weaker matching. Increasing the keyframe count can reduce long-tailed failures by providing more matching opportunities, but the benefit is limited. In some cases, a single keyframe gives a lower median error because fewer noisy correspondences are introduced into Sim(2) fitting, whereas more keyframes increase coverage but may also bring additional outliers.

Orbit trajectories show a different behavior. Their results are independent of keyframe count because orbit-like motion is handled by the BEV branch rather than direct multi-keyframe matching. For orbit motion, a larger window is more stable because VGGT can recover the yaw-to-center rotational structure more consistently, and the resulting BEV representation provides a more reliable source for satellite alignment. In contrast, overly frequent correction with short windows may introduce occasional erroneous anchors that pass validation, leading to lower trajectory accuracy.

The runtime trend is consistent with the pipeline design. Larger windows reduce the number of processed windows and thus the average time per frame, while more keyframes increase the matching cost in the direct branch. Overall, $L=12$ with three keyframes provides a good balance between accuracy and efficiency and is used as the default setting in all experiments.

\subsection{Robustness Analysis}
\label{sec:robustness_analysis}

We evaluate the robustness of ScalLoc to errors in the geometric priors used for
satellite crop prediction. In the direct branch, camera pitch and relative
height are used to estimate the ground footprint and crop the satellite search
region. We add independent per-frame uniform perturbations to these priors: pitch noise
is sampled from $\mathcal{U}(-\delta_\theta,\delta_\theta)$ with
$\delta_\theta \in \{0^\circ,5^\circ,10^\circ\}$, and relative-height noise is
sampled as a proportional perturbation
$\eta_h \sim \mathcal{U}(-\delta_h,\delta_h)$ with
$\delta_h \in \{0\%,10\%,20\%\}$, i.e., the perturbed height is
$(1+\eta_h)h$.

As shown in Table~\ref{tab:robustness_perturbation}, ScalLoc is relatively
robust to height errors. With unchanged pitch, Pos. Acc@20 only decreases from
69.18 to 68.29 when the height perturbation range increases to $\pm20\%$.
Pitch errors have a larger impact: with unchanged height, Pos. Acc@20 decreases
from 69.18 to 63.20 when the pitch perturbation range increases to
$\pm10^\circ$. This is because height errors mainly affect crop scale, whereas
pitch errors also shift the projected ground footprint and reduce
UAV--satellite overlap.

When estimated pitch and height are used for crop prediction, Pos. Acc@20 drops
to 55.71, and ME/MedE increase to $52.03 \pm 128.96$ and 17.19. Nevertheless,
the yaw accuracy remains stable, indicating that absolute positioning is more
sensitive to geometric-prior errors than trajectory orientation.

\label{sec:parameter_analysis}
















\begin{table}[t]
\centering
\caption{Parameter analysis of window size and keyframe count. The best results are highlighted in \textbf{bold}.}
\label{tab:window_keyframe_analysis_by_motion}
\renewcommand{\arraystretch}{1.35}
\setlength{\tabcolsep}{3.8pt}
\resizebox{\columnwidth}{!}{%
\begin{tabular}{ccccccccccc}
\toprule
\multirow{2}{*}{\makecell{\textbf{Window}\\\textbf{Size}}}
& \multirow{2}{*}{\makecell{\textbf{Keyframe}\\\textbf{Count}}}
& \multicolumn{2}{c}{\textbf{Straight}}
& \multicolumn{2}{c}{\textbf{Curve}}
& \multicolumn{2}{c}{\textbf{Orbit}}
& \multicolumn{2}{c}{\textbf{All Frames}}
& \multirow{2}{*}{\makecell{\textbf{Time}\\\textbf{Per Frame}}} \\
\cmidrule(lr){3-4}
\cmidrule(lr){5-6}
\cmidrule(lr){7-8}
\cmidrule(lr){9-10}
& & \textbf{ME} & \textbf{MedE} & \textbf{ME} & \textbf{MedE} & \textbf{ME} & \textbf{MedE} & \textbf{ME} & \textbf{MedE} & \\
\midrule
\multirow{3}{*}{8}
& 1 & 19.77 & 6.93 & 18.09 & \textbf{9.82} & 59.21 & 26.77 & 33.78 & 12.24 & 0.614 \\
& 3 & 16.84 & 8.30 & 13.85 & 11.41 & 59.21 & 26.77 & 33.72 & 13.26 & 0.740 \\
& 5 & 15.14 & 8.11 & \textbf{13.38} & 11.35 & 59.21 & 26.77 & 35.20 & 13.10 & 0.868 \\
\midrule
\multirow{3}{*}{12}
& 1 & 17.21 & \textbf{6.90} & 18.99 & 10.97 & 48.41 & 18.60 & 29.10 & \textbf{11.76} & 0.507 \\
& 3 & 13.99 & 8.69 & 14.90 & 12.53 & 48.41 & 18.60 & 26.97 & 12.66 & 0.561 \\
& 5 & \textbf{13.56} & 8.40 & 14.72 & 12.25 & 48.41 & 18.60 & \textbf{26.77} & 12.43 & 0.628 \\
\midrule
\multirow{3}{*}{16}
& 1 & 22.88 & 7.50 & 18.83 & 14.05 & \textbf{42.15} & \textbf{17.70} & 28.70 & 13.22 & \textbf{0.467} \\
& 3 & 23.14 & 9.76 & 17.69 & 14.31 & \textbf{42.15} & \textbf{17.70} & 28.42 & 13.71 & 0.505 \\
& 5 & 23.11 & 9.43 & 17.60 & 14.50 & \textbf{42.15} & \textbf{17.70} & 28.38 & 13.70 & 0.542 \\
\bottomrule
\end{tabular}%
}
\end{table}

\begin{table}[t]
\centering
\caption{
Robustness analysis under different pitch and relative-height perturbations.
}
\label{tab:robustness_perturbation}

\renewcommand{\arraystretch}{0.95}
\setlength{\tabcolsep}{4pt}

\resizebox{\columnwidth}{!}{%
\begin{tabular}{cccccc}
\toprule
\makecell{\textbf{Pitch}\\\textbf{Perturbation}}
& \makecell{\textbf{Relative Height}\\\textbf{Perturbation}}
& \makecell{\textbf{Pos.}\\\textbf{Acc@20}}
& \makecell{\textbf{Yaw}\\\textbf{Acc@10}}
& \textbf{ME $\pm$ SD}
& \textbf{MedE} \\
\midrule

\multirow[c]{3}{*}{$0.00^\circ$}
& $0.00\%$
& 69.18
& 92.25
& $26.97 \pm 71.67$
& 12.66 \\

& $\pm10.00\%$
& 68.71
& 92.24
& $26.87 \pm 71.44$
& 12.80 \\

& $\pm20.00\%$
& 68.29
& 92.30
& $27.34 \pm 71.65$
& 12.96 \\

\midrule

\multirow[c]{3}{*}{$\pm5.00^\circ$}
& $0.00\%$
& 67.42
& 92.48
& $27.78 \pm 71.91$
& 13.23 \\

& $\pm10.00\%$
& 67.27
& 92.36
& $28.23 \pm 72.58$
& 13.40 \\

& $\pm20.00\%$
& 66.37
& 92.24
& $28.75 \pm 72.64$
& 13.60 \\

\midrule

\multirow[c]{3}{*}{$\pm10.00^\circ$}
& $0.00\%$
& 63.20
& 92.67
& $29.51 \pm 72.23$
& 14.64 \\

& $\pm10.00\%$
& 63.25
& 92.68
& $29.30 \pm 72.17$
& 14.77 \\

& $\pm20.00\%$
& 62.78
& 92.53
& $29.89 \pm 72.20$
& 14.92 \\

\midrule

\multicolumn{2}{c}{VGGT Predictions}
& 55.71
& 90.18
& $52.03 \pm 128.96$
& 17.19 \\

\bottomrule
\end{tabular}%
}
\end{table}

\section{Discussion}
\label{sec:discussion}

The experiments suggest that continuous oblique thermal UAV geo-localization does not necessarily require map-side 3D information. Even when only a geo-referenced 2D satellite map is available, sequential geometric reconstruction can provide useful local structure and relative motion, while repeated satellite-map anchoring can inject absolute geographic constraints and suppress long-horizon drift. This finding indicates that lightweight 2D map references, when coupled with local reconstruction and validated anchoring, can support practical oblique thermal UAV localization in scenarios where DSMs, LoD models, or preconstructed 3D maps are unavailable.

ScalLoc also reveals several limitations. The current motion-adaptive strategy is implemented by a hand-crafted state machine that selects between BEV-based and direct anchoring. Although effective for the studied straight, curve, and orbit patterns, this design may not generalize well to more diverse flight behaviors, scene layouts, or camera configurations. In particular, direct oblique image-to-satellite matching can become unstable under orbit-like motion because rapid viewpoint changes alter the visible footprint and correspondence distribution, making satellite correction less reliable.

Future work should explore a more unified optimization criterion for oblique thermal geo-localization. Instead of selecting anchoring branches with a hand-crafted state machine, a unified formulation could jointly evaluate geometric consistency between predicted camera rays and map-constrained rays, for example by minimizing their angular discrepancy under satellite-map correspondence constraints. Such a criterion may reduce the dependence on motion-specific heuristics, improve robustness under orbit-like viewpoint changes, and provide a more general framework for coupling local reconstruction, cross-view matching, and global map anchoring.

\section{Conclusion}
\label{sec:conclusion}


This paper introduced ScalLoc, a continuous oblique thermal UAV geo-localization framework using only a geo-referenced 2D satellite map. ScalLoc combines windowed reconstruction, track-based pose refinement, motion-adaptive satellite anchoring, and validated propagation to estimate frame-level positions and headings. We also introduced Thermal-Oblique, a benchmark covering diverse scenes, lighting conditions, motion patterns, flight heights, and camera pitch angles. Experiments show that ScalLoc effectively reduces global drift and achieves the lowest all-frame mean position error among the compared methods, demonstrating the feasibility of continuous oblique thermal geo-localization with lightweight 2D satellite maps.

\bibliographystyle{IEEEtran}
\bibliography{reference}

\end{document}